\documentclass[acmsmall]{acmart}

\def\BibTeX{{\rm B\kern-.05em{\sc i\kern-.025em b}\kern-.08emT\kern-.1667em\lower.7ex\hbox{E}\kern-.125emX}}
    
\setcopyright{acmcopyright}
\acmJournal{TIST}
\acmYear{2018}\acmVolume{37}\acmNumber{4}\acmArticle{111}\acmMonth{8}
\acmDOI{10.1145/1122445.1122456}

\begin{document}

%
\title{Combating Fake News: A Survey on Identification and Mitigation Techniques}

%
\author{Karishma Sharma}
\email{krsharma@usc.edu}
\orcid{0000-0001-6825-5876}
\affiliation{%
  \institution{University of Southern California}
}

\author{Feng Qian}
\affiliation{\institution{University of Southern California}}
\email{nickqian@pku.edu.cn}
  
\author{He Jiang}
\affiliation{\institution{University of Southern California}}
\email{jian567@usc.edu}
  
\author{Natali Ruchansky}
\affiliation{\institution{University of Southern California}}
\email{natalir@bu.edu}
  
\author{Ming Zhang}
\affiliation{\institution{Peking University}}
\email{mzhang@net.pku.edu.cn}
  
\author{Yan Liu}
\affiliation{\institution{University of Southern California}}
\email{yanliu.cs@usc.edu}

\renewcommand{\shortauthors}{Sharma, et al.}

%
\begin{abstract}
The proliferation of fake news on social media has opened up new directions of research for timely identification and containment of fake news, and mitigation of its widespread impact on public opinion. While much of the earlier research was focused on identification of fake news based on its contents or by exploiting users' engagements with the news on social media, there has been a rising interest in proactive intervention strategies to counter the spread of misinformation and its impact on society. In this survey, we describe the modern-day problem of fake news and, in particular, highlight the technical challenges associated with it. We discuss existing methods and techniques applicable to both identification and mitigation, with a focus on the significant advances in each method and their advantages and limitations. In addition, research has often been limited by the quality of existing datasets and their specific application contexts. To alleviate this problem, we comprehensively compile and summarize characteristic features of available datasets. Furthermore, we outline new directions of research to facilitate future development of effective and interdisciplinary solutions.
\end{abstract}

%
%

%

%

%
\maketitle

\section{Introduction}\label{sec:intro}

In recent years, the topic of \emph{fake news} has experienced a resurgence of interest in society.  The increased attention stems largely from growing concerns around the widespread impact of fake news on public opinion and events. In January 2017, a spokesman for the German government stated that they ``are dealing with a phenomenon of a dimension that [they] have not seen before'', referring to the proliferation of fake news on social media.\footnote{\url{http://www.theguardian.com/world/2017/jan/09}} Although social media has increased the ease with which real-time information disseminates, its popularity has exacerbated the problem of fake news by expediting the speed and scope at which false information can be spread. \citeauthor{fuller2009decision} \citeyear{fuller2009decision} noted that with the massive growth of online communication, the potential for people to deceive through computer-mediated communication has also grown and such deception can have disastrous and far-reaching results on many areas of our lives, including financial markets \cite{carvalho2011persistent,kogan2018fake} and political events \cite{allcott2017social,aro2016cyberspace}. For instance, \citeauthor{carvalho2011persistent} \citeyear{carvalho2011persistent} noted that a false report of bankruptcy of a United Airlines parent company in 2008 caused the stock price to drop by as much as 76\% in a matter of minutes; although the stock rebounded after the news was identified as false, it closed 11.2\% lower than the previous day and the negative effect persisted for six more days. In terms of political events, an analysis of the US Presidential Election in 2016 by \citeauthor{allcott2017social} \citeyear{allcott2017social} revealed that fake news was widely shared during the three months prior to the election with 30 million total Facebook shares of 115 known pro-Trump fake stories and 7.6 million of 41 known pro-Clinton fake stories. In addition, false stories often emerge surrounding natural disasters such as the Japan earthquake in 2011 \cite{takayasu2015rumor} and Hurricane Sandy in 2012 \cite{gupta2013faking} that are intended to cause increased panic and disorder; or ones surrounding specific individuals and public figures, such as the death hoax of Singapore's first prime minister Lee Kuan Yew \cite{chua2016collective}. Another severe instance of the impact of fake news was the infamous ``Pizzagate" incident wherein physical violence ensued as a result of fake stories circulated online \cite{metaxas2017infamous,fisher2016pizzagate}. It is clear that the magnitude, diversity, and substantial dangers of false information in circulation are a genuine cause for concern.

The topic of fake news has not only received tremendous public attention but has also drawn increasing attention from the academic community. In this regard, there have been attempts to survey and summarize the literature on fake news detection. In one of the first surveys in the area, \citeauthor{shu2017fake} \citeyear{shu2017fake} illustrates an intriguing connection between fake news and social and psychological theories, which suggest that humans tend to seek, consume and believe information that is aligned with their ideological beliefs, which often results in the perception and sharing of false information as true in communities of like-minded people. We note that the past year has witnessed a lot of new work related to fake news, which extends beyond the works surveyed in \cite{shu2017fake}. Another survey by \citeauthor{zubiaga2018detection} \citeyear{zubiaga2018detection} study a related problem of rumor detection, which differs subtly from fake news detection in that it seeks to distinguish between verified and unverified information, wherein the unverified information may turn out to be true or false, or may remain unresolved. The primary objective of this survey was to provide an architecture for developing a rumor classification system. In turn, the work summarizes some of the peripheral issues with identifying and tracking rumors, such as data collection and tools exposed by various social media platforms like Twitter and Facebook for the same, as well as an extensive list of applications and ongoing research projects towards this end. The literature surveyed in \cite{zubiaga2018detection} is specifically restricted to that which is known to have been applied in the context of rumors. However, rumor classification and fake news detection are closely related in characteristics and techniques. In this work, we attempt to bridge the gap by comprehensively summarizing related techniques from closely related contexts. Another survey by \citeauthor{kumar2018false} \citeyear{kumar2018false} addresses a broader scope of false information on the web. In particular, aside from fake news on social media, the survey discusses fake reviews on e-commerce platforms and hoaxes in collaborative platforms such as Wikipedia. The nature and characteristics, and therefore techniques for detection of fake reviews and Wikipedia hoaxes, differ significantly from fake news on social media; for instance, fake reviews are opinion-based where there is no single truth value, information does not propagate over a network, and have specific characteristic such as ratings that are not applicable to fake news. Although these are extremely important topics in their own right, much like the detection of scam email \cite{saberi2007learn}, fake followers \cite{cresci2015fame}, or false web links \cite{lake2014fake}, we chose to keep the discussion focused on fake news and provide detailed discussions of the techniques and their limitations. Moreover, \citeauthor{kumar2018false} \citeyear{kumar2018false} do not discuss some important existing works in fake news detection \cite{ma2016detecting,ma2017detect,ruchansky2017csi,neuralurg2018,wang2017liar} and mitigation such as \cite{pmlr-v70-farajtabar17a,kim2017leveraging,tschiatschek2017detecting} which are important to understand the advances in this domain. In particular, the purpose of our survey is four fold: 

\begin{enumerate}
\item We address both detection and mitigation methods including intervention based techniques, to allow us to understand the problem in an end-to-end manner and go beyond treating the problem as a classification task.
\item We focus on the challenges of developing computational methods to tackle fake news, and present and compare existing methods that have demonstrated significant progresses in dealing with those challenges.
\item Existing surveys lack a comprehensive coverage of available fake news detection datasets. We consolidate a large comprehensive list of datasets and summarize their characteristic features -- we hope this will aid researchers in selecting the right dataset, will facilitate the collection of new datasets, and advance development of methods that can unify different information sources.
\item Additionally, the rapid advancement of research around fake news necessitates the consolidation of recent works and advances and outlining concrete directions for future research.
\end{enumerate}

Fake news identification and mitigation is a critical and socially relevant problem that is also technically challenging for a variety of reasons. In the following sections, we first formally define and characterize fake news and describe the problem and associated challenges. We then present a comprehensive overview of the existing techniques applicable to detection and mitigation, along with a discussion of the key limitations and recent advances in various methods. Further, to facilitate researchers with rigorous evaluation and comparison, we compile a list of existing/available datasets around fake news detection, and summarize their characteristic features. To conclude, we enumerate a list of challenges and open problems that outline promising directions for future research.

\section{Definition, Nature, Associated challenges, Key players} \label{sec:defn}

In this section, we start by defining fake news and describing the nature and characteristics of the problem, as well as the key players involved, and the roles they play in information dissemination, moderation, and consumption. Further, we discuss the main challenges associated with fake news identification and mitigation, and outline the necessary goals of any system that aims to address these challenges in an end-to-end, complete, and practically effective manner.

\subsection{Definition} \label{sec:formal_defn} The usage and meaning of the term \emph{fake news} has evolved over time. A Google Trends Analysis of the term reveals a sudden burst in popularity around the time of the 2016 US presidential election.\footnote{\url{https://trends.google.com/trends/explore?date=2013-12-06\%202018-01-06&geo=US&q=fake\%20news}} Although originally used to reference false and often sensational information disseminated under the guise of news reporting\footnote{as defined in the Collins English Dictionary}, the term has evolved and become synonymous with the spread of \emph{false information} \cite{cooke2017posttruth}. Fake news has generally been defined as ``a news article that is intentionally and verifiably false" \cite{allcott2017social,shu2017fake} or ``information presented as a news story that is factually incorrect and designed to deceive the consumer into believing it is true" \cite{golbeck2018fake}. However the existing definitions are narrow, restricted either by the type of information or the intent of deception, and do not capture the broader scope of the term based on its current usage. Therefore, we define fake news as follows,

\begin{definition}{}
A news article or message published and propagated through media, carrying \emph{false} information regardless the means and motives behind it.
\end{definition}

This definition allows us to capture the different types of fake news identified in \cite{wardle2017fake} which can be differentiated by the means employed to falsify information, such as fabricated content (completely false), misleading content (misleading use of information to frame an issue), imposter content (genuine sources impersonated with false sources), manipulated content (genuine information or imagery manipulated to deceive), false connection (headlines, visuals or captions that do not support the content), and false context (genuine content shared with false contextual information). The definition also allows us to include different types of fake news identified by their motive or intent, such as malicious intent (to hurt or disrepute), profit (for financial gain by increasing views), influence (to manipulate public opinion), sow discord (to create disorder and confusion), passion (to promote ideological biases), amusement (individual entertainment) \cite{zannettou2018web}. We can also subdivide false information by intent as misinformation and disinformation. Misinformation refers to unintentionally spread false information which can be a result of misrepresentation or misunderstanding stemming from cognitive biases or lack of understanding or attention; and disinformation refers to false information created and spread specifically with the intention to deceive \cite{kumar2018false}. Another type of information that might be closely connected to fake news is satire - satire presents stories as news that might be factually incorrect, but the
intent is not to deceive but rather to call out, ridicule, or expose behavior that is shameful, corrupt, or otherwise ``bad" \cite{golbeck2018fake}. The intent behind satire seems legitimate enough to exclude it from the definition, however, \cite{wardle2017fake} does include satire as a type of fake news when there is no intention to cause harm but it has potential to mislead or fool people. Also, \cite{golbeck2018fake} mentions that there is a spectrum from fake to satirical news which they found to be exploited by many fake news sites, which used disclaimers at the bottom of their webpages to suggest they were ``satirical" even when there was nothing satirical about their articles; to protect them from accusations about being fake. Thereby, our definition must include articles that are falsely posed as satire, as well as satirical articles that can potentially mislead, and exclude others that do not fall in this area. Additionally, we disambiguate the terms \textit{hoax} and \textit{rumor} which are closely related to fake news. A hoax is considered to be a false story used to masquerade the truth, and, by the traditional definition, fake news can be seen as a form of hoax usually spread through news outlets \cite{zubiaga2018detection}. The term rumor refers to unsubstantiated claims that are disseminated with the lack of evidence to support them. This makes them very similar to fake news, with the main difference being that they are not necessarily false, and may turn out to be true \cite{zubiaga2018detection}. Rumors originate from unverified sources but may later be verified as true or false, or remain unresolved. Thereby, our definition can be seen as naturally encompassing hoaxes and false rumors.

\subsection{Nature/Characteristics} \label{sec:nature}
The definition of fake news is not the only thing that has changed with time. With the growth of computer-mediated communication through social media, we can see that the nature and characteristics of the problem have also evolved. Hence, we start by reviewing the literature from sociology and psychologically that explain the existence and spread of fake news at both an individual and social level.

\subsubsection{Individual level} \label{sub:individual} The inability of an individual to accurately discern fake from true news leads to the continued sharing and believing of false information in social media.  \citeauthor{YouGov2017} \citeyear{YouGov2017} found in a survey of 1684 British adults who were shown six individual news stories, three of which were true and three of which were fake, only 4 percent were able to identify them all correctly. The inability to discern has been attributed to cognitive abilities and ideological biases. \citeauthor{pennycook2018falls} \citeyear{pennycook2018falls} identified a positive correlation between propensity for analytical thinking and the ability to discern false from true information. In addition, \citeauthor{allcott2017social} \citeyear{allcott2017social} observed that people who spend more time consuming media, people with higher education, and older people had more accurate perceptions of information. The results were statistically significant in a survey of 1208 US adults. Another study examined the impact of cognitive ability on the durability of opinions to find that individuals with lower cognitive ability adjusted their assessments after being told that the information given was incorrect, but not nearly to the same extent as those with higher cognitive ability \cite{roets2017fake}. Besides cognitive abilities, ideological priors play an important role in information consumption. Naive realism (individuals tend to more easily believe information that is aligned with their views), confirmation bias (individuals seek out and prefer to receive information that confirms their existing views), and normative influence theory (individuals choose to share and consume socially safe options as a preference for social acceptance and affirmation) are generally regarded as important factors in the perception and sharing of fake news \cite{shu2017fake}. The survey by \citeauthor{allcott2017social} \citeyear{allcott2017social} also found with statistical significance that people (Democrats and Republicans) are respectively, 17.2 and 14.7 percent more likely to believe ideologically aligned articles than they are to believe nonaligned articles, although the differences in magnitude across the two groups (Democrats and Republicans) are not statistically significant. These individual vulnerabilities have been exploited to successfully disseminate fake information. \citeauthor{higgins2016post} \citeyear{higgins2016post} declared this era as an era of ``post-truth" wherein objective facts are less influential in shaping public opinion than appeals to emotions and personal beliefs\footnote{as defined by the Oxford English Dictionary, 2016}.

\subsubsection{Social level} 
\begin{figure}[t] 
    \centering
    \includegraphics[width=8cm,height=5cm]{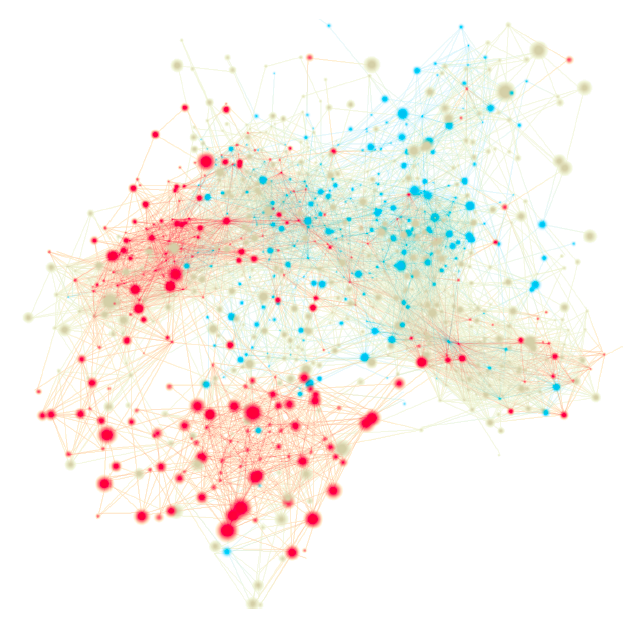}
    \caption{Nodes represent a sample of about 900 Twitter accounts that participated in the conversation about the US Presidential Election between June and mid-September 2016, and edges represent mutual-follower relationships between these accounts. Nodes are sized according to relative PageRank importance in the depicted network, and colored according to inferred political ideology (red: right-leaning, blue: left-leaning, white: unsure) \cite{gillani2018me}. Data source: \cite{gillani2018me}.}
    \label{fig:echo_chamber}
\end{figure}

The nature of social media and collaborative information sharing on online platforms provides an additional dimension to fake news, popularly called the \emph{echo chamber} effect \cite{shu2017fake}. The principles of naive realism, confirmation bias and normative influence theory discussed above in Section \ref{sub:individual} essentially imply the need for individuals to seek, consume and share information that is aligned with their own views and ideologies. As a consequence, individuals tend to form connections with ideologically similar individuals (social homophily), and algorithms tend to personalize recommendations (algorithmic personalization) by recommending content that suits an individual's preferences, as well as by recommending connections to similar individuals to befriend or follow. Both social homophily and algorithmic personalization lead to the formation of echo chambers and filter bubbles, wherein individuals get less exposure to conflicting viewpoints and become isolated in their own information bubble \cite{garimella2017balancing}. The existence of echo chambers can improve the chances of survival and spread of fake news which can be explained by the phenomena of social credibility and frequency heuristic, where social credibility suggests that people's perception of credibility of a piece of information increases if others also perceive it as credible; and frequency heuristic refers to the increase in people's perception of credibility with multiple exposures to the same information \cite{shu2017fake}. \citeauthor{gillani2018me} \citeyear{gillani2018me} examined a network of 1.1M Twitter users who participated in conversations about the US 2016 presidential election between June-September 2016 and observed the existence of echo chambers and polarization of views based on the political orientation (democratic and republic) as visualized in Figure \ref{fig:echo_chamber} that shows the follower relationships (edges) between users marked by the color of their political orientation (which is inferred from a user's profile and tweet contents as per \cite{vijayaraghavan2017twitter}).
\citeauthor{quattrociocchi2016echo} \citeyear{quattrociocchi2016echo} similarly studied polarization by scientific v/s conspiracy narratives in Facebook users and observed that although both types of information were consumed similarly overall, 76.79\% of all users who interacted on scientific pages and 91.53\% of all users who interacted with conspiracy posts had 95\% of their likes on either science or conspiracy posts. In addition, polarized users had a higher number of friends who displayed the same behaviour; and higher edge homogeneity, measuring the similarity between friends in the network, suggested that it is highly unlikely that information would propagate across the different groups.   
\\\\
Fake news diffusion patterns on social media have often been studied to identify the characteristics of fake news that can help differentiate it from true news. Fake news detection works essentially rely on exploiting these differences to classify information based on its veracity. Most existing works primarily target the classification as a binary classification task (fake/true, rumor/non-rumor, hoax/non-hoax) or as a multi-class classification task (true/mostly true/half true/mostly false/false, unverified rumor/true rumor/false rumor/non-rumor). The main difference in different task settings is due to different annotation schemes or applications contexts in different datasets. Usually the datasets are collected from annotated claims on fact-checking websites such as \texttt{PolitiFact, Snopes} and others and therefore reflect the labeling scheme used by the particular fact-checking website/organization. In some instances, credibility scores are provided based on human annotator judgments instead of class labels. A detailed description of different datasets and their annotations is provided in Table \ref{tab:datasets_a} when we summarize and discuss existing datasets in Section \ref{sec:datasets}. In the remainder of this section, we examine different 
characteristics of fake news that are utilized for detection. We can identify three primary characteristics relevant for fake news detection, namely the source/promoters of the information, the information content, and the user responses it receives on social media.

\begin{enumerate}
    \item \textbf{Source/promoters.} \citeauthor{zimdars2016false} \citeyear{zimdars2016false} maintains a list of web addresses of fake news websites; several of which are modified names of true news websites, such as ``abcnews.com.co" and ``washingtonsblog.com". The use of such misleading domain names are a particular characteristic of fake news sources which individuals must learn to recognize and be attentive about. However, there are two caveats to filtering information based solely on these lists. One is that not all articles from these sources are fake and the other is that the list can never be exhaustive. In particular, the use of online resources and social media make it very easy, convenient and inexpensive to create bots and register new accounts or domains \cite{kumar2018false}. Bots are fake or compromised accounts controlled by humans or programs to introduce and promote information on social media. \citeauthor{subrahmanian2016darpa} \citeyear{subrahmanian2016darpa} found a significant overlap between follower and followees of bot accounts that spread false information, which helps to engineer virality and credibility of posts and inflate the social status of the bot accounts. \citeauthor{davis2016botornot} \citeyear{davis2016botornot} found that on Twitter large number of bot accounts were responsible for accelerating the speed of both true and false information roughly equally. \citeauthor{shu2018fakenewsnet} \citeyear{shu2018fakenewsnet} randomly sampled 10,000 users who posted fake and real news on Twitter and used the bot detection algorithm by \cite{davis2016botornot} to find that almost 22\% of users involved in fake news are bots, while only around 9\% of users are predicted as bot users for real news. They also investigated the temporal patterns of account activities by capturing the relationship between number of posts on Twitter at different times and days of the week for posts related to fake and true news; and observed that periods of high posting activity included odd hours when people are generally inactive, suggesting the existence of social bots \cite{shu2018fakenewsnet}. The temporal patterns observed by \citeauthor{shu2018fakenewsnet} \citeyear{shu2018fakenewsnet} are shown in Figure \ref{fig:heatmap_bot}. Lastly, analyzing how source and promoters of fake news operate over the web across multiple online platforms, \citeauthor{zannettou2018web} \citeyear{zannettou2018web} found that false information is more likely to spread across platforms (18\% appearing on multiple platforms) compared to true information (11\%), with Reddit to Twitter to 4chan being the most common direction of information flow.\\

    \item \textbf{Information content.} The content of the information being spread is primarily what needs to be classified as true or fake. \citeauthor{horne2017just} \citeyear{horne2017just} identified certain characteristics that differentiate fake news contents from true news contents by studying the textual characteristics of articles from various fake news websites and contrasting them with articles from reputed journalistic websites. Their findings suggest that titles of fake news articles are longer, have more capitalized words, and use fewer stop words; and the body content of fake news articles are shorter, repetitive, have fewer nouns, and analytical and technical words. \citeauthor{perez2017automatic} \citeyear{perez2017automatic} found that fake news articles contained more social words, with more verbs and temporal words suggesting that the text tends to be focused on the present and future, instead of being more objective and factual. Other textual cues include self-reference, negation statements, complaints, and generalizing items \cite{volkova2018misleading}. \citeauthor{newman2003lying} \citeyear{newman2003lying} analysis revealed that deceptive stories had lower cognitive complexity, fewer exclusive words, more negative emotion words, and more motion (action) words. \citeauthor{silverman2015lies} \citeyear{silverman2015lies} observed that 13\% of 1600 news articles had incoherent headlines and content, and used declarative headlines paired with article bodies which are skeptical about the claim in the headline \cite{kumar2018false}. \\
   
    
     \begin{figure}[t] 
    \centering
    \includegraphics[width=14cm,height=4cm]{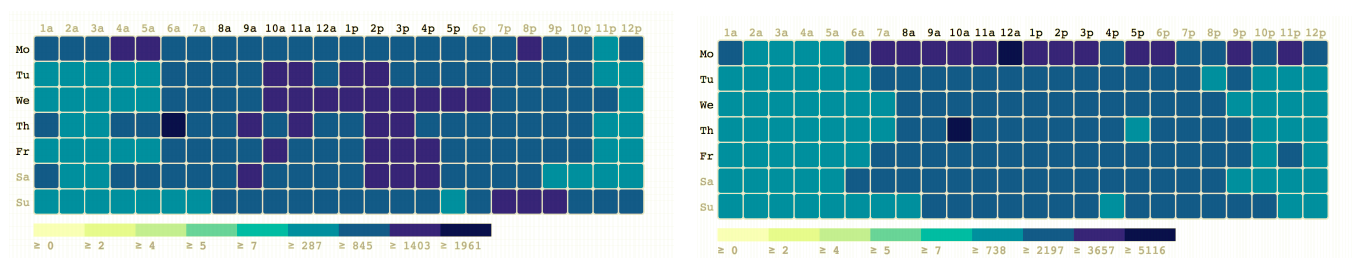}
    \caption{The heatmap of the day of week vs hour of tweets posted related to fake news (left) and real news (right). Darker colors indicate greater posting activity in a particular time slot \cite{shu2018fakenewsnet}. Data source: \cite{shu2018fakenewsnet}.}
    \label{fig:heatmap_bot}
    \end{figure}
    \item \textbf{User responses.} User responses on social media provide auxiliary information that is extremely beneficial for fake news detection. They are known to provide stronger signals for detection than the information content, mainly because user responses and propagation patterns are harder to manipulate than the information content, and oftentimes contain obvious information about the veracity \cite{shu2018fakenewsnet}. This secondary information in the form of user engagements (likes, shares, replies or comments) contains rich information captured in the \emph{propagation structure} (tree) which indicates the path of information flow, temporal information in \emph{timestamps} of engagements, textual information in \emph{user comments}, and \emph{user profile} information by the user involved in the engagement. \citeauthor{zubiaga2018detection} \citeyear{zubiaga2018detection} note that it is possible to differentiate user responses by their stance, wherein the most commonly used categorization uses the following four types, namely, supporting, denying, querying and commenting (which can be neutral or unrelated). They also observe that the nature of user responses varies depending on the stages of propagation, where, in the case of rumors, it was observed that majority of the users support true rumors and a higher number deny false rumors when the entire life cycle of the rumor was considered; whereas by studying only the early reactions to rumors, it was found that users showed a tendency to support rumors independent of their veracity, which is suggestive of the fact that users have problems determining veracity in the early stages. \citeauthor{neuralurg2018} \citeyear{neuralurg2018} similarly found that fake news tends to receive more negative and questioning responses than true news. Another analysis of true and fake news captured the variation of sentiments in user replies, wherein it was observed that sentiment in replies to true news tended more towards neutral, in contrast to that for fake news that tended more towards negative sentiments \cite{shu2018fakenewsnet}. 
    \citeauthor{friggeri2014rumor} \citeyear{friggeri2014rumor} also found from an analysis of user responses on Facebook, that user responses change once false information is debunked, with a 4.4 times increase in deletion probability, even in early stages of the propagation.
\end{enumerate}

\subsection{Information Exchange Process}
There are several different entities (individuals and organizations) that are \emph{simultaneously} at play when it comes to the dissemination, moderation and consumption of fake news through social media, which makes the problem of identification and mitigation more complex and involved. We discuss each part of the information exchange process.

\subsubsection{Dissemination} A noticeable shift in information dissemination channels from traditional forms of journalism to online social media has been observed \cite{paine2015next,perrin2015social}. In a survey of 3,000 journalists, 20\%, responded that they thought that social media spelled the death of journalism \cite{paine2015next}. Social media sites have become the popular form of dissemination due to growing ease of access and popularity of computer-mediated communication. Fake news can have a larger impact through social media due to the large scale and reach of social media; and the ability to collaboratively share content \cite{hermida2012tweets}.

\subsubsection{Moderation} While in traditional forms of journalism the responsibility of content creation rests with the journalist and the reporting organization, moderation in social media varies substantially. In 2017, Germany passed the NetzDG (Network Enforcement) Act\footnote{https://www.bbc.com/news/technology-42510868} to enforce removal of fake news within 24 hours (or up to a week depending on the complexity) by social media platforms with more than two million users. In UK, the parliament launched an investigation into how fake news is threatening modern democracy \cite{harriss2017online}. Nevertheless, the question of distribution of responsibility remains unresolved. A recent survey by \citeauthor{barthel2016many} \citeyear{barthel2016many} determined that Americans collectively assign a fairly high and equal amount (45\%) of responsibility to the government/politicians, social media platforms/search engines, and lastly, members of the public; and specifically 15\% hold all three responsible, while 27\% hold two and 31\% hold one of three responsible. \cite{barthel2016many}. 

\subsubsection{Consumption} Information is primarily consumed by the general public or society which is a growing body of social media users. In a 2018 survey on social media usage with 2002 US adults surveyed, \cite{pew2018social} established that 68\% Americans use of Facebook, 73\% use Youtube and between 20-40\% use other social media platforms such as Twitter and Instagram, and the numbers have grown almost ten-fold since 2005 \cite{perrin2015social}. \citeauthor{pew2018social} \citeyear{pew2018social} also established that 74\% of the Facebook users use the site daily, with 51\% using it several times a day; and a large fraction of frequent users lie in the 18-29 range. This growth in information consumption through social media adds to the risks of fake news causing wide-spread damage.
 
\subsection{Key Players}
We now consider a more subtle aspect that usually characterizes fake news, that is, the intent to deceive. In this light, we discuss the different roles that entities (individuals and organizations) play when it comes to dealing with fake news.
 
\subsubsection{Adversary} Malicious individuals and organizations with a political or social agenda often pose as ordinary social media users using social bots \cite{bessi2016social}
or actual accounts and can act as the source as well as promoters of fake news. Such accounts are known to also indulge in group behavior wherein groups of such accounts coordinate and share the same set of fake news articles \cite{ruchansky2017csi}. The general characteristics of sources and promoters of fake news have been discussed in Section \ref{sec:nature}.

\subsubsection{Fact-checker} In an attempt to combat the growing amount of false information, various \emph{fact-checking} organizations, such as \texttt{Snopes} and \texttt{Politifact}, have been initiated to expose or confirm news stories. While these organizations are based on ``fact-checking journalism", that relies on human verification, more desirable automated technological solutions have been proposed by technological companies like \texttt{Factmata} which aim to provide fake news detection solutions to businesses and consumers and assign credibility scores to web content using artificial intelligence. Various other automated solutions in the form of plug-ins and applications such as \texttt{BS-Detector} and \texttt{CrossCheck} provide similar automated fact-checking services. An exhaustive list of fact-checking applications is provided in \cite{zubiaga2018detection}.

\subsubsection{Susceptible} Fake news affects a wide range of individuals and organizations based on the motive. We summarized different motives or intents behind the spread of fake news in Section \ref{sec:formal_defn}. For instance, reputable institutions and individuals might be susceptible to the attacks of fake news that is intended to sway public opinion about them, such as what was witnessed during the US 2016 presidential election \cite{allcott2017social}. Other consequences can even prove to be an increased risk to the entire world. \citeauthor{roozenbeek2018fake} \citeyear{roozenbeek2018fake} noted that false information discrediting the seriousness of global warming can affect people's perception of climate change, posing a great risk to society and the world at large.

\subsection{Challenges} The nature of the problem presents several challenges, which we summarize as follows.
\subsubsection{High stakes and multiple players} The World Economic Forum (2013) has ranked the spread of false information as one of the ``top risks" the world is facing today. Moreover, the involvement of multiple entities and technological platforms increases the difficulty of studying and designing computational, technological and business strategies, without compromising rapid and collaborative access to high quality information.

\subsubsection{Adversarial intent} Malicious intent in content design and promotion increases the complexity of the problem. The content is designed to not only make it harder for humans to identify fake news by exploiting their cognitive abilities, emotions and ideological biases as discussed in Section \ref{sub:individual}, but also to make it more challenging for computational methods to detect fake news. \citeauthor{shu2018fakenewsnet} \citeyear{shu2018fakenewsnet} evaluated the performance of several different methods on two datasets from PolitiFact and GossipCop and reported a maximum detection accuracy of 69\% and 79.6\% respectively, even when using both article contents and social context i.e. user responses to the article on Twitter.

\subsubsection{Public susceptibility and lack of awareness} To raise public awareness, numerous articles and blogs have been written that provide tips on differentiating truth from falsehood.  For example, award-wining journalist Laura McClure highlighted five important questions to ask yourself when trying to determine whether a news article is true or fake. In one of the questions, McClure asked the reader to consider how an article makes them feel, citing that fake news is often ``designed to make you feel strong emotions''.\footnote{\url{http://blog.ed.ted.com/2017/01/12/how-to-tell-fake-news-from-real-news/}}  Articles such as McClure's are informative and effective for individuals; however, they do not provide a scalable and systematic solution to the problem.

\subsubsection{Propagation dynamics} The dynamic nature of the process of fake news propagation through social media further complicates matters. False information can easily reach and impact a large number of users in short time \cite{neuralurg2018,friggeri2014rumor}. \citeauthor{friggeri2014rumor} \citeyear{friggeri2014rumor} studied rumor cascades on Facebook and found that information is readily and rapidly transmitted, even when it is of dubious veracity. Fact-checking organizations like \texttt{Snopes} and \texttt{Politifact} cannot keep up with the propagation dynamics as they require human verification which inhibits a timely and cost-effective response \cite{ruchansky2017csi,kim2017leveraging}. 

\subsubsection{Constant change} Fast-paced developments in the world pose additional challenges to knowledge base systems which need to dynamically retrieve and update their state based on newly emerging facts \cite{potthast2017stylometric}. Fact-checking is ultimately essential for reliably identifying fake news. For example, while scandals are not uncommon among celebrities, when a new scandal about a celebrity comes out, without enough extra knowledge, it is very difficult to tell whether it is fake or not. \citeauthor{potthast2017stylometric} \citeyear{potthast2017stylometric} note that style-based fake news detection i.e. differentiating fake and true news by an analysis of writing styles are simply alternatives used due to the unresolved challenges in automating fact-checking from knowledge bases. 

\subsection{Requirements/Goals}

Existing works have demonstrated significant progresses towards alleviating some of the challenges in fake news detection and mitigation. However, there is still room for improvement before the problem can be addressed in more effective ways. We suggest few requirements that are of interest in developing solutions to tackle fake news.

\subsubsection{Balancing aggressive and non-aggressive moderation} Identification and mitigation techniques that require very aggressive moderation by social media platforms can hurt the social media platform \cite{pmlr-v70-farajtabar17a}. Thereby, it seems necessary to design strategies that still effectively mitigate the problem of fake news, but without restricting rapid access to high quality information, and collaborative information sharing. We believe that it is also important that the moderation strategies do not further introduce more confusion and distrust into the environment. Fake news has already resulted in people becoming skeptical of even true information, which hurts the value of social media as a platform for information sharing. A survey of 1002 Us adults \cite{barthel2016many} found that (64\%) people say that fake news has caused a great deal of confusion about the basic facts of current issues and events, 24\% say it has caused some confusion and 11\% say not much/no confusion.

\subsubsection{End-to-end solutions} Reliable and timely detection of fake news should be accompanied by computational methods to intervene and prevent further spread of news that is confirmed as fake. In addition, it would be desirable to design interventions that can provide quantifiable measures of the impact of the intervention effort in terms of the exposures to fake and true news \cite{pmlr-v70-farajtabar17a}.

\subsubsection{Balancing timeliness v/s detection accuracy} Early detection and mitigation are critical goals of any effective system. However, the available information for detection increases as time progresses, with only the content of the article being available at the start, followed by increasing user responses as propagation continues \cite{neuralurg2018}. Most existing methods either rely on content only or on user responses only, or do not utilize responses incrementally. Detection systems must aim to utilize incrementally available information to trade-off confidence in detection accuracy v/s timeliness of the detection and mitigation effort. 

\subsubsection{Prioritization and cost-effectiveness} The ability to optimally decide which contents to fact-check at what time, can equip the system in providing better responses by being able to quickly remove false information that can have a potentially larger and faster impact than those that might have a negligible or slower impact if allowed to propagate further in time \cite{papanastasiou2017fake,kim2017leveraging}. Also, human involvement in fact-checking increases not only the delay but also the cost of intervention, which necessitates the need for prioritization of information to manually fact-check, until reliable automated methods can be sought \cite{papanastasiou2017fake}.

\subsubsection{Robustness, scalability and interpretability} The high stakes and consequences of fake news necessitates the need for reliability in detection. Mistakenly removing true information from the platform, or not detecting and removing potentially viral false information would become problematic in practice. To move from manual and semi-automated solutions to fully automated ones will not be possible without robust and also interpretable predictions.  \citeauthor{popat2018credeye} \citeyear{popat2018credeye} emphasized transparency and interpretability in credibility assessment systems and built CredEye which verifies an input claim and provides the probability of truth and falsehood along with extracted evidence (from fact-checking or trusted news websites) with words that support the claim highlighted in green, refuting the claim highlighted in red and words overlapping with the claim provided in yellow, and the intensity of colors reflects the
word's importance for the assessment (based on feature weights from the learned classifier).

\subsubsection{Evolution and up-to-date fact-checking} The adversarial nature of fake news necessitates that the system should be able to dynamically adapt to the changing strategies of adversarial opponents. Specifically, adversaries create new accounts and throw-away accounts to promote fake news and avoid detection \cite{kumar2018false}, as can be seen in an instance of the US election when potentially large number of fake accounts were created to influence the election and were found to initiate coordinated attacks with specific political agendas\footnote{https://www.nytimes.com/2017/09/07/us/politics/russia-facebook-twitter-election.html}. In addition, increased sophistication in adversarial techniques are speculated in terms of both fake content creation \cite{kumar2018false}, as well as in strategies to promote fake content. \citeauthor{ruchansky2017csi} \citeyear{ruchansky2017csi} observed from behaviours of suspicious users on Twitter, that these users had more similar engagement patterns towards true and fake news than on Weibo, which could demonstrate an increased sophistication in fake content promotion on Twitter.

\section{Overview of methods}

We divide the existing work into three types. The first type is fake news identification using content-based methods, that classify news based on the content of the information to be verified. The second type is identification using feedback-based methods, that classify news based on the user responses it receives on social media. Lastly, the third type is intervention based solutions, that provide computational solutions for \emph{actively} identifying and containing the spread of false information, and methods to mitigate the impact from exposures to false information. Each category is further divided based on the type of existing methods, as shown in Table \ref{tab:org}.

\begin{table*}
\caption{Categorization of existing methods}
\label{tab:org}
\begin{tabular}{ lll }
\toprule
Content-Based Identification & Feedback-Based Identification & Intervention-Based Solutions \\
\midrule
Cue and feature methods & Hand-crafted features  & Mitigation strategies \\  
Linguistic analysis methods & Propagation pattern analysis & Identification strategies  \\
Deep learning content-based & Temporal pattern analysis & \\
 & Response text analysis & \\
 & Response user analysis & \\
 \bottomrule
\end{tabular}
\end{table*}

\section{Content-based identification}

In this section, we give an overview of content based approaches to fake news detection. The underlying basis of content based detection is that textual content in fake news differs from that in true news in some quantifiable way. The use of language cues to determine veracity was first motivated by work in applied psychology for evaluation of eyewitness testimonies \cite{undeutsch1984courtroom}. Language cues can be exploited with traditional hand-engineered feature based methods, linguistics based methods, and more advanced deep learning methods that bypass feature engineering for content analysis.

\subsection{Cue and feature based methods}
Cue and feature based methods can be employed to distinguish fake news contents from true news contents by designing a set of linguistic cues that are informative of the content veracity. Several different cue sets have been proposed in literature.

\subsubsection{Scientific Content Analysis (SCAN)}
\label{sec:text} One of the earliest works on examining the use of linguistic cues for fake news detection was by \citeauthor{driscoll1994validity} \citeyear{driscoll1994validity}, that studied transcripts or written statements made by individuals in a criminal investigation. In order to determine the credibility of the information given by suspects in such statements, they utilized an approach called Scientific Content Analysis (SCAN), which was proposed by the polygraph examiner \citeauthor{sapir1987lsi} \citeyear{sapir1987lsi}, based on his experience with subjects of polygraph examinations. SCAN consists of cues related to deception detection. The cues include content and structural criteria such as lack of connection between paragraphs, lack of conviction or memory, denial of allegations, missing and out of order information, use of emotive words, objective or subjective words, pronouns, first person, singular, and past tense verbs. While the examination by \citeauthor{driscoll1994validity} \citeyear{driscoll1994validity} revealed positive results in differentiating true from false statements using SCAN, later works have found them to be ineffective based on more rigorous evaluations, finding no significant differences between true and fabricated statements concerning these criteria \cite{nahari2012does,bogaard2016scientific}.

\paragraph{Limitation} Although intuitive, SCAN is set of subjective criteria lacking enough supporting evidence; and moreover, the approach requires the use of trained professionals to analyze the statements for veracity, making it difficult to automate.

\subsubsection{Linguistic-Based Cue Set (LBC)} In an attempt to reduce human involvement in deception detection, one of the pioneering automated text analysis methods was by \citeauthor{fuller2009decision} \citeyear{fuller2009decision}. \citeauthor{fuller2009decision} \citeyear{fuller2009decision} consolidated several linguistic-based cues and previously proposed cue sets. The first cue set included was the Zhou/Burgoon set \cite{zhou2004automating} comprising 14 linguistic based cues that were found effective for deception detection included percentage of first person singular and plural pronouns in the text, average word length, verb quantity, sensory ratio, spatial and temporal ratio, and imagery. The second set of cues was derived from deception constructs drawn from deception theories \cite{buller1996interpersonal,buller1996testing} that included sentence and word quantity, activation, certainty terms, generalizing terms, imagery, and verb quantity. The third cue set was a comprehensive set of 31 cues created by including the first two cue sets along with additional Linguistic Inquiry and Word Count (LIWC) \cite{pennebaker2001linguistic} based cues utilized by previous studies \cite{newman2003lying,bond2005language}, and included lexical diversity, modal verbs, passive verbs, emotiveness, exclusive terms, and redundancy. The final cue set was a refined cue set determined by feature selection to identify the most important of the 31 cues in the comprehensive set. To determine the relative importance of cues, \citeauthor{fuller2009decision} \citeyear{fuller2009decision} utilized three different classifiers, that is neural network, decision tree, and logistic regression, on statements of witnesses in official investigations; wherein the input feature vector to the classifier consisted of the normalized frequency of cues appearing in the text. The eight cues obtained through feature selection were third person pronouns, content word diversity, exclusive terms, lexical diversity, modifiers, sentence quantity, verb quantity and word quantity.

\paragraph{Limitation} The drawback of using linguistic-based cue sets is the lack of generalizability across topics, languages and domains. \citeauthor{ali2008language} \citeyear{ali2008language} showed that a linguistic cue set designed for one situation may not be suitable for another situation due to language variations e.g. a cue set designed for accounting~\cite{larcker2012detecting} or police interrogation~\cite{porter1996language} may differ significantly.



\subsubsection{Other variants} Later works have explored refined hand-crafted cue sets more closely targeted towards the problem of fake news detection.  \citeauthor{rubin2016fake} \citeyear{rubin2016fake} analyzed several text analysis based features, such as the number of punctuation marks and the sentiment of the text. \citeauthor{zhao2015enquiring} \citeyear{zhao2015enquiring} proposed different regular expressions to capture enquiry and correction patterns in posts on social media, which are indicative of rumors such as ``is (that/this/it) true" which capture enquiry and ``(that/this/it) is not true" which capture correction. They also included some platform specific features such as counts of `hashtags', and `mentions' in posts on Twitter, as well as the ratio of enquiry or correcting posts in a cluster of posts having high textual similarity i.e. posts discussing similar content. Others works that have proposed cue sets designed specifically for certain types of social media platforms include \cite{castillo2011information,gupta2014tweetcred,morris2012tweeting} for Twitter, \cite{kumar2016disinformation} for Wikipedia, \cite{chen2015misleading} for click-bait websites.

\paragraph{Limitation} Exhaustive enumeration of regular expression patterns is non-trivial and requires significant effort. Furthermore, identifying relevant platform specific features for a large variety of social media platforms is also challenging and reduces applicability of the method across platforms and domains.
\\\\
\textbf{Limitations of cue and feature based methods.} To summarize the limitations, variations in linguistic cues implies that a new cue set must be designed for a new situation, making it hard to generalize cue and feature engineering methods across topics and domains. And such approaches thereby involve more human involvement in the process to design, evaluate and utilize these cues for detection.




















\subsection{Linguistic analysis based methods} 
While manual cue selection is intuitive and interpretable, it is often specific to the setting being considered and does not generalize easily to other settings. In an attempt to make cue-based models more general, methods based on linguistic analysis were proposed. Linguistic analysis based methods, like cue-based methods, can be applied to distinguish fake from true news by exploiting differences in writing style, language and sentiment. Such methods do not require task-specific, hand-engineered cue sets and rely on automatically extracting linguistic features from the text. We discuss three linguistic analysis based methods that are applied to fake news detection.

\subsubsection{N-gram Approach} \label{sec:linguistic_datasets} The most effective linguistic analysis method applied to fake news detection is the \emph{n-gram} approach ~\cite{ott2011finding,ott2013negative,mihalcea2009lie}. n-grams are sequences of $n$ contiguous words in a text, constituting words (unigrams) and phrases (bigrams, trigrams) and are widely used in language modeling and text analysis.

\paragraph{Approach} \citeauthor{mihalcea2009lie} \citeyear{mihalcea2009lie} proposed the use of n-grams for lie detection. They constructed datasets using crowd-sourcing which constituted statements of people lying about their beliefs on topics such as abortion and death penalty / lying about their feelings on friendship. \citeauthor{mihalcea2009lie} \citeyear{mihalcea2009lie} wanted to determine how the texts differed and whether n-grams analysis was enough to differentiate lies from the truth. They trained Naive Bayes and Support Vector Machine (SVM) classifiers with inputs being the term frequency vectors of n-grams in the texts, after tokenization and stemming but without stop word removal. Interestingly enough the classification accuracy was about 70\% in identifying people's lies about their beliefs, and 75\% in identifying lies about their feelings. A fine-grained analysis of word usage revealed that in all the deceptive texts, connections to the self (``I, friends, self") were lacking and other human-related word classes (``you, other, humans") were dominant, indicative of the speaker's discomfort in identifying themselves with the lying statements. Also, it was found that words related to ``certainty" were dominant in deceptive texts, which is probably explained by the need for the speaker to explicitly use truth-related words as a means to make their false statements more believable. \cite{ott2011finding,ott2013negative} provided similar n-gram classification analysis for deceptive reviews created by crowd-sourced workers on Amazon Mechanical Turk, who were asked to create fake positive reviews about hotels \cite{ott2011finding} and fake negative reviews about hotels \cite{ott2013negative}. Their analysis revealed that fake reviews contained less spatial words (location, floor, small), because the individual had not actually experienced the hotel and had less spatial detail available for the review. They also found that positive sentiment words were exaggerated in positive fake reviews compared to their true counterparts. A similar exaggeration was seen in negative sentiment words in fake negative reviews.

\paragraph{Limitation} Being a simplified approach, using n-grams alone cannot entirely capture finer-grained linguistic information present in the writing styles of fake news.


\subsubsection{Part-of-Speech Tags} Apart from word based features such as n-grams, syntactic features such as Part-of-Speech (POS) tags are also exploited to capture linguistic characteristics of texts. POS tags are obtained by tagging each word in a sentence according to its syntactic function, such as nouns, pronouns, adjectives; and several works have found the frequency distribution of POS tags to be closely linked to the genre of the text being considered, for example, medical consultations, committee meetings, and sermons, each have their own distinctive pattern~\cite{biber1999longman,rayson2001grammatical}. \citeauthor{ott2011finding} \citeyear{ott2011finding} examined whether this variation in POS tag distribution also exists with respect to text veracity. They trained a SVM classifier using relative POS tag frequencies of texts as features on a dataset containing fake reviews. \citeauthor{ott2011finding} \citeyear{ott2011finding} obtained better classification performance with the n-grams approach, but nevertheless found that the POS tag approach is a strong baseline outperforming the best human judge. A qualitative analysis showed that the weights learned by the classifier are largely in agreement with the findings of existing theories on deceptive writing such as \cite{rayson2001grammatical} which suggests connections of deceptive opinions to imaginative writing comprising more verbs, adverbs, pronouns and pre-determiners, and truthful opinions to informative writing comprising more nouns, adjectives, prepositions, determiners and coordinating conjunctions.

\paragraph{Limitation} The sole use of POS tags provides syntactic information alone and is weaker than word based approaches that capture more information, inclusive of writing styles such as emotiveness inferred from words like excited, terrible, etc.

\subsubsection{Probabilistic Context Free Grammar} Later work has considered deeper syntactic features derived from Probabilistic Context Free Grammars (PCFG) trees \cite{johnson1998pcfg}. Context Free Grammar (CFG) tree represents the grammatical structure of a sentence with the terminal nodes representing words and intermediate nodes representing syntactic constituents such as verb, noun phrase, etc. Depending on language construction ambiguities, a sentence can have multiple syntactic representations. PCFG enables disambiguation by associating a probability with each tree, where the probability of a tree is the product of the probabilities of all production rules in the tree. A production rule is represented as follows, $A \rightarrow \alpha$ where $A \in V$ and $\alpha \in (V \cup T)^*$; $V$ being the set of intermediate nodes and $T$ being the set of terminal nodes. The production rule probability is statistically derived from the corpus.

\paragraph{Approach} \citeauthor{feng2012syntactic} \citeyear{feng2012syntactic} examined the use of PCFG to encode deeper syntactic features for deception detection. In particular, they proposed four variants when encoding production rules as features. The first variant includes only those production rules from the data that do not contain terminal nodes. The second variant includes all production rules derived from the dataset. The third and fourth variant modifies the production rules to include grandparent nodes (i.e. parent of node $A$ in rule $A \rightarrow \alpha$), with and without rules with terminal nodes respectively. The feature vector is constructed using the tf-idf counts (normalized frequency) of production rules in the text. \citeauthor{feng2012syntactic} \citeyear{feng2012syntactic} trained an SVM classifier, and found PCFG features used with n-gram features to be more beneficial than POS tags with n-grams in classifying fake texts from hotel reviews and lie detection datasets \cite{ott2011finding,mihalcea2009lie}. 
\citeauthor{feng2012syntactic} \citeyear{feng2012syntactic} noted that of the four variants of production rules, the ones with the terminal nodes included were more powerful since they contained word based information in addition to syntax information. A qualitative analysis of the most discriminative syntactic constituents based on the classifier weights indicated the use of certain syntactic constituents such as indirect enquiry (WHADVP), verb phrases (VP), and subordinating conjunction clauses (SBAR) to be more frequent and important in fake texts as compared to true texts.

\paragraph{Limitation} PCFG can be used to extract syntactic features from a sentence but is not powerful enough to capture context sensitive information from across sentences, thereby limiting its effectiveness in classification of longer fake news articles or texts.
\\\\
\textbf{Limitations of linguistic analysis based methods.} Even with word based n-gram features combined with deeper syntactic features from PCFG trees, linguistic analysis methods although better than cue-based methods, still do not fully extract and exploit the rich semantic and syntactic information in the content. n-gram approach is simple and cannot model more complex contextual dependencies in the text. Syntactic features used alone are less powerful than word based n-grams, and a naive combination of the two cannot capture their complex interdependence.

\subsection{Deep learning content-based methods}
\label{sec:deep_text}
Deep learning methods alleviate the shortcomings of linguistic analysis based methods by automatic feature extraction, being able to extract both simple features and more complex features that are difficult to specify. Deep learning based methods have demonstrated significant advances in text classification and analysis \cite{kim2014convolutional,kalchbrenner2014convolutional,yih2014semantic} and are powerful methods for feature extraction and classification with their ability to capture complex patterns relevant to the task.

\subsubsection{Convolutional Neural Networks} 
\begin{figure}[t]
\centering
\includegraphics[width=10cm,height=5cm]{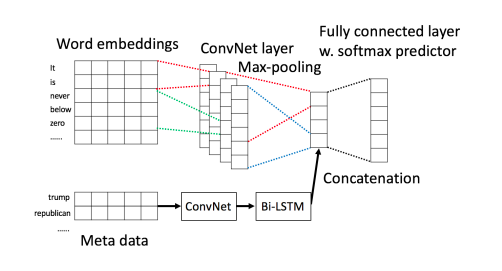}
\caption{CNN architecture used for fake news detection \cite{wang2017liar}. Data source: \cite{wang2017liar}.}
\label{fig:cnn}
\end{figure}

Convolutional neural networks (CNN) \cite{lecun1998gradient} are generally used in natural language processing tasks such as semantic parsing \cite{yih2014semantic} and text classification \cite{kim2014convolutional}. \citeauthor{wang2017liar} \citeyear{wang2017liar} proposed the use of convolutional neural networks for content-based fake news detection. \citeauthor{wang2017liar} \citeyear{wang2017liar} collected a dataset of short statements labeled based on the degree of falsehood by \texttt{PolitiFact}, a reputed fact-checking organization. Figure \ref{fig:cnn} shows the model architecture proposed by \cite{wang2017liar}. The model takes two inputs, the statement text and the speaker metadata information such as political orientation, home state, etc. available in the dataset. The text inputs are processed by a word embedding layer to obtain continuous low dimensional representations for each word in the text sequence. The output of this layer is processed by convolutional and max-pooling layers that generate the extracted feature representation. Similarly, the speaker metadata is also processed similarly by a different embedding and convolutional layer and a bidirectional LSTM \cite{hochreiter1997long} layer to generate its final extracted feature representation. The two representations are concatenated and fed to the classifier trained end-to-end with the other layers. \citeauthor{wang2017liar} \citeyear{wang2017liar} utilized pre-trained word2vec embeddings \cite{mikolov2013efficient} to warm start the text embeddings. These word embeddings are known to capture useful properties of word co-occurrences and contextual properties, and semantic relationships between words, trained on large corpora of unlabeled texts. \citeauthor{wang2017liar} \citeyear{wang2017liar} observed better detection accuracy with the deep learning based CNN model compared to using SVM and logistic regression and also found the inclusion of speaker metadata beneficial.
\citeauthor{neuralurg2018} \citeyear{neuralurg2018} also similarly demonstrated improved CNN performance over linguistic analysis based methods such as LIWC, POS and n-gram approach when classifying a collection of news articles as fake or true. In addition, to handle longer article texts, \citeauthor{neuralurg2018} \citeyear{neuralurg2018} suggested a variant of the CNN architecture called Two-Level Convolutional Neural Network (TCNN) which first takes an average of word embedding vectors for words in a sentence to generate sentence representations, and then represents articles as a sequence of sentence representations provided as input to the convolutional and pooling layers. \citeauthor{neuralurg2018} \citeyear{neuralurg2018} found the TCNN variant to be more effective than CNN in classifying the articles.

\subsubsection{Other Variants} \label{sec:rnn} Recurrent neural network (RNN) \cite{rumelhart1988learning} based architectures are also proposed for fake news detection. RNNs process the word embeddings in the text sequentially, one word/token at a time, utilizing at each step the information from the current word to update its hidden state which has aggregated information about the previous words. The final hidden state is generally taken as the feature representation extracted by the RNN for the given input sequence. A specific variant called Long Short-Term Memory (LSTM) \cite{hochreiter1997long}, which alleviates some of the training difficulties in RNN, is often used due to the its ability to effectively capture long range dependencies in the text, and has been applied to fake news detection, similar to the use of convolutional neural networks, in several works \cite{volkova2017separating,rashkin2017truth}; whereas in another variant, LSTM has been applied to both, article headline and article text (body), in an attempt to classify the level of disagreement between the two for deception detection \cite{chopra2017towards}.
\\\\
\textbf{Limitations of deep learning based methods.} Even with sophisticated feature extraction of deep learning methods, fake news detection remains to be a challenge, primarily because the content is crafted to resemble the truth in order to deceive readers; and without fact-checking or additional information, it is often hard to determine veracity by text analysis alone. A recent evaluation \cite{shu2018fakenewsnet} benchmarking different methods on datasets of political statements and celebrity gossip news also confirms relatively low classification accuracy of 63\% and 70\% on the two datasets using purely content-based classification with CNN.

\section{Feedback-based identification}

In this section, we present and discuss feedback based approaches for fake news detection. In content based approaches, the text of an article is regarded as the primary source of information. However, rich secondary information in the form of user responses and comments on articles and patterns of news propagation through social media, can likely be more informative than article contents which are crafted to avoid detection. These secondary information sources form the basis of the works discussed in this section.

\subsection{Hand-engineered features}
Hand-engineered features proposed for fake news detection from feedback signals include various types of features such as, propagation pattern features, temporal pattern features, text-related and user-related features. \citeauthor{castillo2011information} \citeyear{castillo2011information} designed a feature set to include user-based features (e.g. registration age and number of followers), text-based features (e.g. the proportion of tweets that have a mention `@' ), and propagation based features (e.g. the depth of the re-tweet tree) and used a decision tree model to classify news as fake using the designed feature vector. Variants of the above include other network-based features that are slightly extended or tailored to a particular context, such as the inclusion of temporal features \cite{kwon2017rumor}, or geographic location and type of device (mobile or PC) from which the response was sent \cite{yang2012automatic}.

\paragraph{Limitation} Feature engineering allows us to incorporate diverse kinds of information, which is useful in this case. However, hand-engineered features are limited in terms of the generality and complexity of the feature space being captured.

\subsection{Propagation pattern analysis}
\label{sec:propagation}
True and fake news spread through social media in the form of shares and re-shares of the source and shared posts, resulting in a diffusion cascade or tree, with the source post at the root. The path of re-shares and other propagation dynamics of fake and true news contents are utilized for fake news detection. In this section, we discuss works that specifically utilize propagation structures and patterns for fake news detection.

\subsubsection{Propagation tree kernels}
Propagation tree kernel methods exploit the differences in propagation patterns of two diffusion cascades (trees) as features for detection. \citeauthor{ma2017detect} \citeyear{ma2017detect} proposed to compare the similarity between propagation trees using \emph{tree kernels} that were applied to syntactic modeling tasks in \cite{zhang2008exploring,moschitti2006efficient,collins2002convolution}.

\paragraph{Approach} \citeauthor{ma2017detect} \citeyear{ma2017detect} defined a tree kernel which is utilized to compute the similarity between two trees as follows. Each node of the tree corresponds to a specific user engagement with the article; and is associated with the timestamp of the engagement, textual information (user's comment/reply) and the user's metadata. These attributes form the feature set of the node. The similarity between two nodes (one from each propagation tree) is a function of the similarity between the node features and similarity over subtrees, computed recursively. \citeauthor{ma2017detect} \citeyear{ma2017detect} designed the tree kernel with different measures of similarity for different features. For textual features, they utilized Jaccard similarity  as a measure of similarity over n-grams in the textual information $J(c_i, c_j)$. For the temporal similarity, they considered the absolute difference of the time lags $t = |t_i - t_j|$, where the time lag is basically the relative difference of timestamp of the engagement and the timestamp of the source post. For user metadata features, they considered the L2-norm (euclidean distance) between the user metadata feature vectors $ \mathcal{E}(u_i, u_j)$. Combining the three metrics of similarity, the similarity between two nodes $v_i$ and $v_j$ is defined as follows, \[ f(v_i, v_j) = e^{-|t_i - t_j|}(\alpha \mathcal{E}(u_i, u_j) + (1-\alpha) J(c_i, c_j)), \] where $v_i$ is a node from one propagation tree and $v_j$ from the other, $\alpha$ is a constant, $u_i, u_j$ are user metadata vectors, and $c_i, c_j$ are the textual information n-gram vectors, $t_i,t_j$ are the time lags of the engagements. Using the defined kernel, \citeauthor{ma2017detect} \citeyear{ma2017detect} computed the similarity between every pair of diffusion cascades in the dataset, and used SVM with the defined tree kernel for classification, observing improved detection performance over methods that do not utilize propagation patterns. \citeauthor{wu2015false} \citeyear{wu2015false} similarly defined a random walk graph kernel \cite{kang2012fast} to calculate similarity between different propagation trees. The designed kernel function is composed of the random walk graph kernel over propagation trees and the standard RBF (radial basis function) kernel over other features, paired with SVM, similar to \cite{ma2017detect}. 

\paragraph{Limitation} Propagation tree kernel methods are computationally intensive (requiring the computation of pairwise similarities over all trees), which can prohibit their large scale applicability to the task of fake news detection.

\subsubsection{Propagation tree neural networks} The limitations of propagation tree kernel methods have been addressed by more recent works \cite{ma2018rumor} using deep learning methods. The advances proposed in \citeauthor{ma2018rumor} \citeyear{ma2018rumor}, not only significantly reduce the computation costs and time, but also manage to improve over the detection accuracy obtained by the tree kernel methods \cite{ma2017detect}. \citeauthor{ma2018rumor} \citeyear{ma2018rumor} proposed the use of recursive neural networks, generally used in syntactic and semantic parsing \cite{socher2011parsing,socher2012semantic,socher2013recursive} to extract propagation features from diffusion cascades.

\begin{figure}[t] 
\centering
    \includegraphics[width=\linewidth,height=5cm]{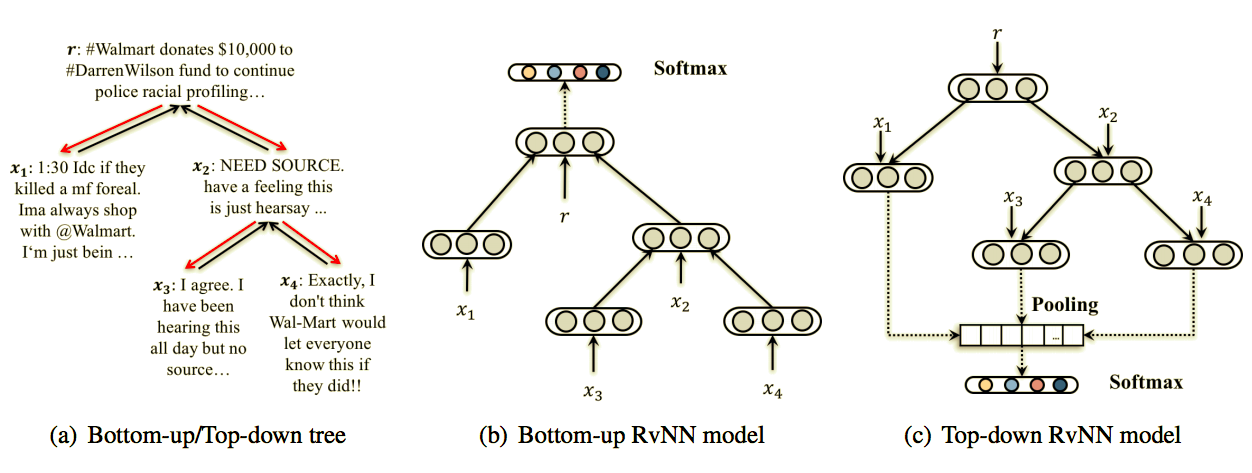}
    	\caption{Tree structured recursive neural network architectures considered in \cite{ma2018rumor}. Data source: \cite{ma2018rumor}.}
    \label{fig:ma2018rumor}
\end{figure}

\paragraph{Approach} \citeauthor{ma2018rumor} \citeyear{ma2018rumor} design two recursive neural network models based on bottom-up and top-down tree-structured recursive neural networks. Similar to recurrent neural networks (RNN) mentioned in Section \ref{sec:rnn}, that sequentially process each timestep (word) in the input sequence while aggregating information from previous timesteps (words), the recursive tree neural network sequentially processes each node in the tree by recursively traversing the tree structure in either the top-down or bottom-up manner based on the design. In the bottom-up approach, information is aggregated starting from leaf nodes and propagated upwards, such that the final feature representation obtained at the root is representative of the complete propagation tree, and is used for classification. The weights in the classification layer (output layer with softmax activation) and the parameters of the recursive network are trained end-to-end based on the classification loss. In the top-down approach, information flows over paths in the tree and the resulting representations at the leaves are combined to generate the extracted feature representation of the tree that is used by the classification layer. Figure \ref{fig:ma2018rumor} shows the proposed network architectures. \citeauthor{ma2018rumor} \citeyear{ma2018rumor} find that both architectures outperform propagation tree kernels and other baselines that do not utilize tree structured information in user engagements. They also find that the top-down approach performs better, which they believe is because it assumes the direction of information flow through the diffusion cascade.

\paragraph{Limitation} The sequential recursive nature of computations impacts training speed. In addition, since information will be aggregated and compressed over many nodes, in large cascades it might potentially not retain enough information from all nodes to obtain a good representation for the tree.

\subsubsection{Propagation process modeling}
Alternatively, extensions to models originating from epidemiology that model the spread of diseases \cite{newman2002spread,kimura2009efficient} have also been applied to characterize the propagation of fake news. \citeauthor{jin2013epidemiological} \citeyear{jin2013epidemiological} proposed a mathematical model called SEIZ to capture how people share news on social media. \citeauthor{jin2013epidemiological} \citeyear{jin2013epidemiological} consider a set of eight news stories spanning multiple topics such as politics, terrorist events, etc. propagating on Twitter, of which four are true news, and the rest are rumors, and examine if the proposed model can characterize the studied cascades. 

\paragraph{Approach} The proposed model SEIZ is defined by dividing users into four partitions: susceptible users (`S') that have not heard about the news yet, exposed users (`E') who have received the news and will share it in the future (with some delay), infected users (`I') that have shared it, and skeptic users (`Z') that have heard about the news but do not share the news. A user can transition from one partition to another with a certain probability and at a certain rate, as defined by the model. 
The parameters of the model $\beta,\rho, b, \epsilon$ are the rates of transitions between different partitions, whereas $p, l$ are transition probabilities (for instance, probability of a `S' user moving to `Z' is $l$ and to `E' is $1-l$, with rate $b$). It is assumed that each partition has an associated size at time $t$, but the total size of all partitions combined remains constant over time (i.e. users only transit from one partition to another and the propagation dynamics are controlled by the parameters of the model). The size of a partition at a given time is obtained by solving a set of differential equations based on the initial size of partitions and the rate constants and probabilities, which form the parameters of the model. The parameters are estimated using nonlinear least squares fit on an observed cascade corresponding to true or fake news. Each step of the fitting process involves iteratively refining a set of parameter values by numerically solving the system of differential equations to minimize the difference between observed shares and the size of partition `I' based on the model $|I(t) - \textrm{observed shares (t)}|$. \citeauthor{jin2013epidemiological} \citeyear{jin2013epidemiological} quantify a ratio ($R_{si}$) as the ratio of influx into E from S to the outflow from E to I, and suggest that it can potentially be used to detect rumors, where they found higher $R_{SI}$ is suggestive of true news based on the studied cascades. The ratio is defined as follows,
\[R_{SI} = \frac{(1-p \beta) + (1-l)b}{p +\epsilon} \]
where $p$, $\rho$, $\beta$, $b$, $l$ and $\epsilon$ refer to the model parameters mentioned earlier.

\paragraph{Limitation} The model parameters are not user specific and the same rate constants and probabilities are assumed for all users. Moreover, the model has not been specifically tested for fake news detection to validate its empirical success or computational feasibility.

\subsection{Temporal pattern analysis}
\label{sec:temporal}
Differences in temporal dynamics of user engagements with articles can be leveraged for fake news detection. The length of time and the rate and intervals at which true and fake news spreads can differ. Extracting intricate interactions and variations along time explicitly is the focus of the work presented in this section.

\subsubsection{Temporal variation features}
Temporal modeling methods are applied to sequentially ordered responses $T_i=\{t_1, t_2 \cdots t_n\}$, in which each post $t_i$ has a time stamp at which the interaction or user engagement with the article is recorded. In the simplest form, temporal features can be incorporated by taking both, features in each time interval and the variation between features of two temporally adjacent intervals \cite{ma2015detect}. In particular, \citeauthor{ma2015detect} \citeyear{ma2015detect} consider three types of features, that is, text features (such as percentage of posts with exclamation or question marks, percentage of posts with hashtags or user mentions, average number of positive/negative emoticons), user features (such as percentage of verified users, average number of followers, registration age) and propagation features (such as average number of reshares, average number of comments). \citeauthor{ma2015detect} \citeyear{ma2015detect} divide every sequence of ordered posts into a fixed number of time intervals and each post in the sequence falls within some interval. Based on that, they compute the features for each time interval from aggregation of posts up to that interval. In addition, the variation of features between two adjoining time intervals is considered as the differences in the feature values divided by the interval length. The final feature vector is a concatenation of features from each interval together with the temporal variation features, which is paired with an SVM classifier in \cite{ma2015detect}. \citeauthor{ma2015detect} \citeyear{ma2015detect} observe reasonable performance improvements over methods that do not consider temporal patterns and rely only on the overall statistics such as total number of reshares, or time length of propagation, and ignore their variations over time.

\paragraph{Limitation} One shortcoming of such a primitive feature variation model is that it requires hand-crafted features which are further restricted to numerical features over which temporal variations can be considered.


\subsubsection{Temporal pattern with recurrent neural networks}\label{sec:temporalrnn} 
Intricate temporal variation differences between true and fake news can be automatically extracted with deep learning based methods. Recurrent neural networks (RNN) are effective in time series modeling. The approach used in general for application to fake news detection is to divide the sequentially ordered engagements into discrete time intervals. Each interval is represented by a set of features. The ordered sequence of feature vectors is provided as input to the recurrent neural network. Each timestep is processed by the network by building on information processed in previous timesteps. 

\paragraph{Approach} \citeauthor{ruchansky2017csi} \citeyear{ruchansky2017csi} partition a given sequence of engagements using discrete time intervals at specific granularity. Timestamps of engagements are considered relative to the first engagement in the sequence. Each interval is represented by the aggregated features of engagements in that interval. Specifically, the feature vector has the following form, where $\eta$ is the number of engagements in interval $t$ , $\triangle t$ is the time between the current and previous non-empty interval, $x_u$ is the average of user-features over users that engaged with the article during $t$, and $x_{\tau}$ is the textual content in engagements during $t$:
\[ x_t = (\eta, \triangle t, x_u, x_{\tau})\]
Together, $\eta$ and $\triangle t$ provide a general measure of the frequency and distribution of the response an article received. The textual features $x_{\tau}$ can be generated in different ways. \citeauthor{ruchansky2017csi} \citeyear{ruchansky2017csi} in particular, chose to learn $x_{\tau}$ from the raw text features using the doc2vec~\cite{le2014distributed} model. \citeauthor{ruchansky2017csi} \citeyear{ruchansky2017csi}  trained an LSTM recurrent neural network \cite{hochreiter1997long} with the sequence of feature vectors $x_t$ as input for classification of fake news based on the user engagements. They used two datasets, one with engagements on Twitter, and the other on Weibo. Both datasets did not contain article contents and the classification is done solely using the user responses (engagements) available in the dataset, using the proposed LSTM model. The datasets also does not contain any propagation tree information and therefore, the engagements of a given article can only be ordered temporally. The model architecture proposed by \citeauthor{ruchansky2017csi} \citeyear{ruchansky2017csi} also contains a second component. The first component we discussed here is to capture the temporal and textual patterns in user responses. The second component is to separately capture user characteristics and group behaviours and is explained in \ref{sec:user_group}. Another earlier work by \citeauthor{ma2016detecting} \citeyear{ma2016detecting} also examined the use of recurrent neural networks to capture temporal patterns from engagements. The architecture by \citeauthor{ma2016detecting} \citeyear{ma2016detecting} does not include the second component capturing user characteristics. The feature representation chosen in the two works differs slightly, in that \citeauthor{ma2016detecting} \citeyear{ma2016detecting} uses tf-idf vectors as text features, and temporal differences are not provided as features as in \citeauthor{ruchansky2017csi} \citeyear{ruchansky2017csi}, but are instead captured by sampling engagements at regular intervals from the time series. The slight differences in feature representations chosen by the two do not affect model performance. \citeauthor{ma2016detecting} \citeyear{ma2016detecting} also examined the performance of other RNN variants besides LSTM, such as Gated Recurrent Unit (GRU) \cite{cho2014learning}, and a two layered GRU network, and found the latter to have slightly better performance out of all three variants.

\paragraph{Limitation} Cascades with too many user responses, would possibly require some sampling or pruning scheme to be efficiently represented, which may result in distortion of the temporal patterns.

\subsection{Response text analysis}

The text in user responses can be quite informative towards fake news detection. Even without access to the article content, response text can provide at least some insights about the article content. It is also likely that fake or dubious articles will receive more negative and questioning responses which can be leveraged for detection. In this section, we discuss methods that specifically exploit textual content in user responses. 

\begin{figure}[t]
\centering
\vspace*{0.1in}
\includegraphics[width=12cm]
{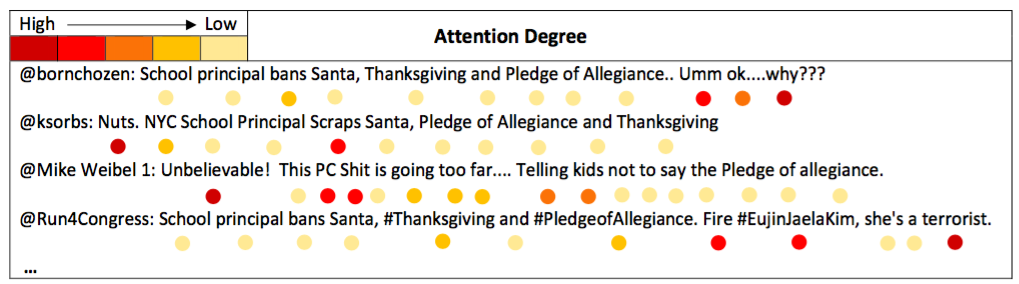}
\vspace*{0.1in}
\caption{Attention weights learned over different words in the sentences. Darker color is indicative of greater attention, making the corresponding words more relevant to the fake news detection task \cite{chen2017call}. Data source: \cite{chen2017call}.}
\label{fig:attention}
\end{figure}

\subsubsection{Deep attention}
Work discussed in Section \ref{sec:temporalrnn} utilized tf-idf features and doc2vec word embeddings to represent textual response features. Another variant of that work proposed by \citeauthor{chen2017call} \citeyear{chen2017call} focuses specifically on the extracted textual information by adding an attention mechanism into the LSTM architecture to capture representative words of fake news and are able to depict which portions of the text can be indicative of truth and deceit; an example is shown in Figure~\ref{fig:attention}, wherein words marked in darker red color appeared to be more important for the fake news detection task, and vice versa. Although adding attention was not found to specifically improve detection performance, it provides some interpretability in the extracted features.

\paragraph{Limitation} The deep attention used is nice for providing insights into interpretability but is not seen to improve detection accuracy in \citeauthor{chen2017call} \citeyear{chen2017call}, thereby is more useful for qualitative studies on fake news rather than detection.

\subsubsection{User response generation} Fake news detection research is generally split between content based and feedback based identification, i.e. detection using either article text or using secondary information from user engagements with the article. \citeauthor{neuralurg2018} \citeyear{neuralurg2018} provides a new approach to integrate the two sources of information directly targeted towards \emph{early fake news detection}. User responses to articles can only be available after the news has propagated sufficiently in order to collect enough number of user responses to apply detection methods. This can lead to increasing costs and is averse to the timeliness of detection. The article text on the other hand is available from the beginning. The early detection setting considered by \citeauthor{neuralurg2018} \citeyear{neuralurg2018} assumes that only the article text is available at the time of detection, which is similar to detection using content based methods.
\begin{figure}[t] 
    \includegraphics[width=5cm,height=4cm]{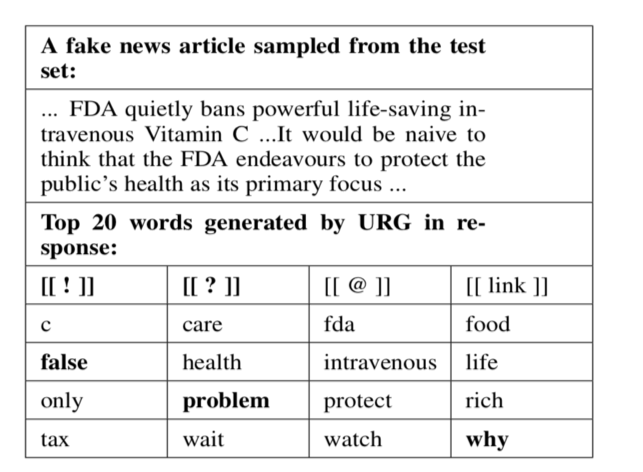} 
    \hspace{10px}
    \includegraphics[width=7cm,height=4cm]{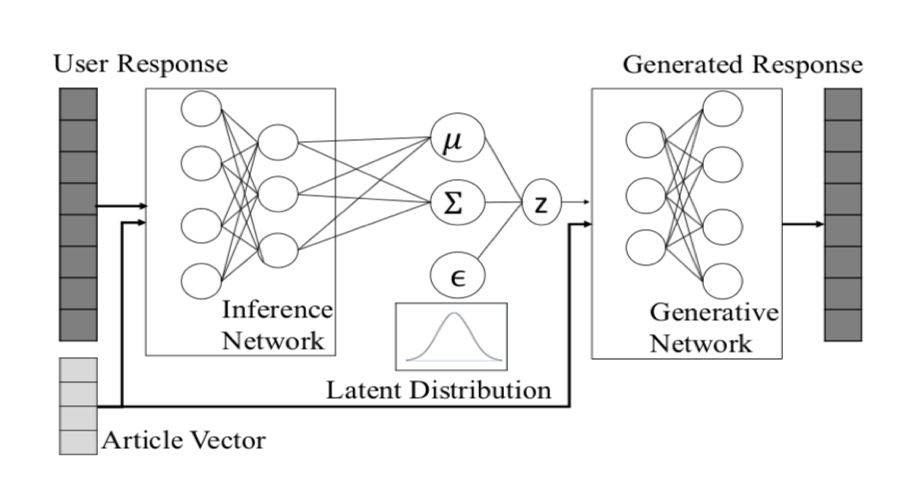}
    \caption{Top 20 words from the generated response samples to a fake news article (left) and the user response generator network (right) \cite{neuralurg2018}. Data source: \cite{neuralurg2018}.}
    \label{urg}
\end{figure}

\paragraph{Approach} Unlike content based methods, \citeauthor{neuralurg2018} \citeyear{neuralurg2018} propose a new strategy based on probabilistic generative modeling to still leverage rich secondary information without having access to responses for the article being classified. The strategy is to instead utilize historically collected articles and associated user responses to learn a probability distribution over user responses conditioned on the article, which can then be used to generate sampled responses to the article being classified. The conditional generative model (i.e. user response generator URG) is shown in Figure \ref{urg}. Article features are extracted using a convolutional neural network (CNN) trained on article word features. The obtained article representation along with the user response words are used to then train the URG. During detection, the article representation is first extracted using the CNN and used to condition the URG to generate sample responses to the article by sampling from the learned distribution. At this time, the URG receives only the article representation as an input since there are no user responses collected at the time of detection, and relies on the learned distribution to generate sample responses. The final classification is done using the concatenated features of the extracted article representation from the CNN and the generated samples of user responses obtained from the URG conditioned on the article representation. The approach is found to have improved detection performance over using only the extracted article representation for detection.

\paragraph{Limitation} \citeauthor{neuralurg2018} \citeyear{neuralurg2018} only utilize a simplistic bag-of-words representation (vector of term frequency of words) of the textual features of the user response, that are provided as input to the URG (generative model). Such representations might not capture more intricate syntactic and semantic information available in the word sequences of texts.

\subsubsection{Stance detection} User responses (comments) have also been utilized to improve detection performance by explicitly taking into account their stance towards the content being discussed. It is generally considered that each user response has one of four stances, namely, support, deny, query or comment. Recent works have considered multi-task learning  approaches to jointly provide stance detection and veracity classification in order to improve classification accuracy by utilizing the interdependence in the two tasks \cite{kochkina18all,ma2018detect}. Both these works proposed multi-task learning architectures based on recurrent neural networks with shared and task specific parameters. 

\begin{figure}[t] 
    \includegraphics[width=14cm,height=5cm]{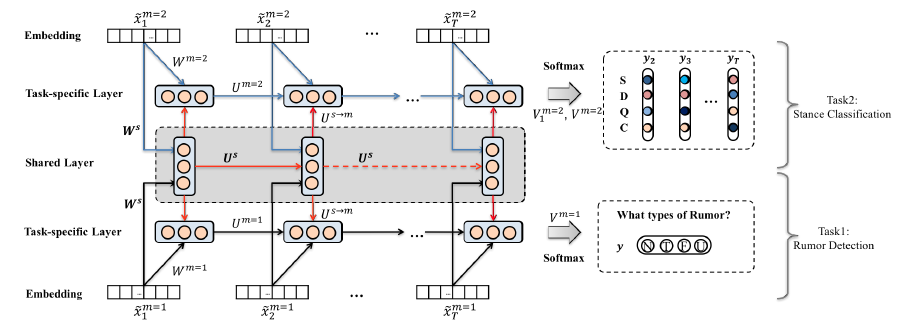}
    \caption{Neural multi-task learning for stance detection and veracity classification jointly \cite{ma2018detect}. Data source: \cite{ma2018detect}.}
    \label{stance}
\end{figure}

\paragraph{Approach} \citeauthor{ma2018detect} \citeyear{ma2018detect} combine the tasks of stance detection and veracity classification where each task is assigned a shared GRU layer and a task-specific GRU layer. The purpose of the shared GRU layer is to capture patterns common to both tasks, whereas task specific GRU enables the capture of patterns that are more important to one task than to the other. For instance, veracity classification relies more strongly on patterns directly conveying veracity such as ``true" and ``false", whereas patterns like ``believe", ``don't think" can be more directed towards stance detection. The architecture proposed by \citeauthor{ma2018detect} \citeyear{ma2018detect} is depicted in Figure \ref{stance}. The input is a sequence of posts (user responses) represented by text (tf-idf) features. The input sequence of vectors are converted to low dimensional representations using a shared embedding layer and task-specific embedding layers and provided as input sequences to the respective shared and task-specific GRU layers. To enhance the interaction between the task-specific layer and shared layer, the hidden state at each time step of the task-specific GRUs are additionally made dependent on the hidden state from the shared GRU layer as shown in Figure \ref{stance}. In the veracity classification task, the hidden state at the last time step of the task-specific GRU is utilized for classification using a fully-connected output layer with softmax activation specific to the veracity classification task. In the stance detection task, every post in the sequence is classified using a fully-connected output layer with softmax activation specific to the stance detection task. \citeauthor{ma2018detect} \citeyear{ma2018detect} found that joint learning of stance detection and veracity classification tasks improves the performance of individual tasks. Utilizing shared and task-specific parameters were observed to be more beneficial than using only the shared parameters without the task-specific layer.  \citeauthor{kochkina18all} \citeyear{kochkina18all} proposed a similar approach with shared and task-specific LSTM layers for rumor detection, stance classification and rumor veracity classification jointly (rumor detection was specified as classifying the sequence of posts as responses to a rumor or non-rumor; and the rumor veracity classification task was specified as classifying the rumors as true, false or remains unverified). In addition, each input sequence is considered as a sequence of posts along a particular branch of the propagation tree, instead of considering all posts by their temporal ordering regardless of the propagation tree as in \citeauthor{ma2018detect} \citeyear{ma2018detect}. 

\paragraph{Limitation} The process of annotating the stance of each user response, which is required as training data can be crowd-sourced but is still a time intensive process.

\subsection{Response user analysis}
\label{sec:user}
User reputation has long been a considered factor in many Internet services such as \emph{Amazon, LinkedIn, TripAdvisor}, and other e-commerce applications\footnote{\url{http://www.cnn.com/2000/TECH/computing/11/07/suing.ebay.idg/}}, and on content sharing platforms like \emph{Wikipedia, StackOverflow, Quora}. In the case of fake news propagation, users can be involved as sources or promoters of misinformation. In this section, we discuss efforts made towards characterization of users posting and propagating news over social media.

\subsubsection{User features} User features can be obtained based on two types of information. One is the features extracted from user profiles on social networks. The second is features extracted based on user behaviors from content sharing and response patterns of the user. For the first type of features, hand-crafted user features like registration age of users, number of followers, and the like, are leveraged in \citeauthor{castillo2011information} \citeyear{castillo2011information} along with other textual and propagation features for fake news detection. Another way of obtaining user representations is proposed by \citeauthor{wu2018tracing} \citeyear{wu2018tracing} who construct user representations using network embedding approaches on the social network graph. For the second type of features, \citeauthor{tacchini2017some} \citeyear{tacchini2017some} construct a feature vector consisting of the set of users who \emph{liked} (responded) to the source article or post, and used these feature vectors to train a logistic regression model for classification of the source article or post as fake. The model essentially tries to capture patterns in user behaviors across articles for classification. Similarly, \citeauthor{qazvinian2011rumor} \citeyear{qazvinian2011rumor} compute probability distributions over users involved in fake and true posts from collected user engagements in order to model user behavior patterns. 

\paragraph{Limitation} Hand engineered user features can be restrictive to one social media platform and not generalize to others. Individual user profile and behaviour should be complemented by analysis of user group behaviours.

\subsubsection{User group analysis} \label{sec:user_group}
Most previous work treat individual user information separately, without considering group behaviors of users. \citeauthor{ruchansky2017csi} \citeyear{ruchansky2017csi} noted that there are groups of users who share the same set of fake articles. According to the analytical study of information sharing on social media~\cite{del2016spreading}, ``users tend to aggregate in communities of interest'', and ``users mostly tend to select and share content according to a specific narrative and to ignore the rest.'' Group behaviors of users are therefore beneficial for generalization of features across users and in identifying anomalous user groups responsible for the promotion of fake news. 

\begin{figure}[t]
\centering
\includegraphics[width=\textwidth,height=5cm]
{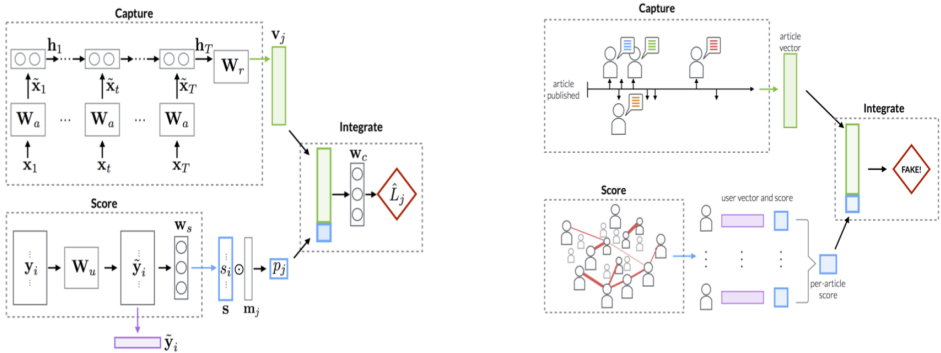}
\caption{Temporal pattern analysis and user group behavior analysis \cite{ruchansky2017csi}. Data source: \cite{ruchansky2017csi}.}
\label{fig:group_anomaly}
\end{figure}

\paragraph{Approach} To detect group behaviours, \citeauthor{ruchansky2017csi} \citeyear{ruchansky2017csi} constructed a weighted user-user graph where an edge between users denotes the number of articles that both have co-engaged with. The user feature vector is constructed using singular value decomposition on the user-user graph i.e. the user co-engagements matrix. The user features constructed this way are lower dimensional distributed representation of users reflective of user group engagements and behaviors. The architecture of the detection model is shown in Figure \ref{fig:group_anomaly}. The capture component is used to extract temporal information from user engagements as explained in Section \ref{sec:temporalrnn}. The score component is used to learn a learn a suspiciousness score for each user and takes user feature vectors as input. It is observed that the method is able to generate meaningful suspiciousness scores over users, which are positively correlated with the engagements of the users in true and fake article cascades. In addition, the detection performance is observed to improve due to the integration the score component capturing user group behaviors, into the overall architecture. Additionally, the benefit of the above method is that the user feature extraction does not require labeled cascades and we can extract user features from co-engagement information derived from unlabeled cascades itself. Labeled cascades are expensive to obtain.

\paragraph{Limitation} The main limitation is that in cases where new fake accounts and social bots are created specifically to spread fake news, there might be no collected user behaviors for such accounts and no direct way to generalize extracted user features to these accounts.
\\\\
\textbf{Limitations of existing feedback based methods.} One significant drawback of existing feedback based classification methods discussed above is that the models are trained on a snapshot of the user responses generally collected after or towards the end of the propagation process, when sufficient responses are available. This explains a drop in performance on early detection using the trained models when fewer responses are available. The methods do not have the ability to update their state based on incrementally available user responses.

\label{sec:network}

\section{Intervention-based solutions}
The scope of the methods discussed in content based and feedback based identification is limited to classifying news from a snapshot of features extracted from social media. For practical applications, we need techniques that can dynamically interpret and update the choice of actions for combating fake news based on real time content propagation dynamics. We discuss the works that provide such computational methods and algorithms in this section.

 



\subsection{Mitigation strategies}
Exposure to fake news can lead to a massive shift in public opinion as noted in Section \ref{sec:intro}. We discuss several strategies aimed at reversing this effect by strategically introducing true news into the social media platforms such that users can be exposed to the truth and the impacts of fake news on user opinion can be mitigated. Computational methods designed for this purpose, require us to first consider some of the widely used information diffusion models. There are several well known models used to represent information diffusion in social networks, such as the \textit{Independent Cascade} (IC) and \textit{Linear Threshold} (LT) model \cite{kempe2003maximizing} and point process models such as \textit{Hawkes Process} model \cite{zhou2013learning}. The IC model represents each edge with a parameter $p_{u,v}$ denoting the strength of influence of user $u$ on user $v$. The diffusion process starts with a set of seed nodes assumed to be activated at the first timestep. At each time step of the diffusion process, a node $u$ activated at time step $t$, independently makes a single activation attempt on each inactive neighbor $v$. The activation succeeds with probability $p_{u,v}$ and a node once activated remains activated throughout the diffusion process. In the LT model the edge parameters represent a weight of influence and additionally each user has a separate (uniformly and independently) random threshold parameter and the sum of weights of incoming edges should be less than one. Accordingly, activation of a user occurs if the weights of all its activated neighbors exceed its threshold at a given timestep of the diffusion process. Point process models are defined differently from the above two, by defining an intensity function which we discuss later in Section \ref{sec:hawkes}.  

\subsubsection{Decontamination} \label{sec:decon}
\citeauthor{nguyen2012containment} \citeyear{nguyen2012containment} proposed a strategy to decontaminate users that have been exposed to fake news. The diffusion process is modeled using the \textit{Independent Cascade} or \textit{Linear Threshold} model \cite{kempe2003maximizing} described earlier. A greedy algorithm is designed to select the best set of seed users from which to start the diffusion process for true news, so that at least a $\beta$-fraction of the users can be decontaminated. The algorithm is a simple greedy algorithm that iteratively selects the next best user to include into the seed set based on the marginal gains obtained by the inclusion of the user (i.e. the number of users that will be activated or reached by the true news in expectation, if the seed set did additionally include the chosen user). The iterative selection of seed users is continued till the objective of decontaminating $\beta$-fraction of the contaminated users in expectation can be achieved. \citeauthor{nguyen2012containment} \citeyear{nguyen2012containment} show that this greedy algorithm has a $(1-1/e)$ approximation ratio with respect to the optimal seed set that should be selected.

\paragraph{Limitation} One drawback of the proposed approach is that it is suggested as a corrective measure \emph{after} the spread of fake news. Secondly, the number of seeds is not fixed, and if the damage done is large the cost of correction by activating or incentivizing more seed users to start the true news cascade might be prohibitive.

\subsubsection{Competing cascades} \label{sec:compcas}
Several other works propose a more interesting intervention strategy based on competing cascades; wherein a true news cascade is introduced to compete with the fake news cascade as the fake news originates and begins to propagate through the network, rather than \emph{after} its propagation. In particular, \citeauthor{budak2011limiting} \citeyear{budak2011limiting} and \citeauthor{he2012influence} \citeyear{he2012influence} formulated an \emph{influence blocking maximization} objective, as finding an optimal strategy to disseminate true news in the presence of a misinformation cascade by strategically selecting $k$ seed users, with the objective of minimizing the number of users who at the end of the diffusion are activated by the fake news campaign instead of the true news one. The model assumes that a user once activated by either the fake or true cascade, remains activated under that cascade. Each edge contains a separate set of diffusion model parameters one for true news and the other for fake. It is reasonable to model the two types of diffusion processes separately, as the strength of influence under true news might not be the same when sharing fake news. \citeauthor{budak2011limiting} \citeyear{budak2011limiting} showed that a similar greedy algorithm as discussed in Section \ref{sec:decon} achieves a $(1-1/e)$ approximation ratio to the optimal under the restriction that the true news influence probabilities are all 1 or exactly equal to the negative influence probabilities. \citeauthor{he2012influence} \citeyear{he2012influence} similarly showed that under the \emph{Linear Threshold} model, the greedy algorithm ratio is $(1-1/e)$ without requiring restrictions on the influence weights.

\paragraph{Limitation} The selection of seed users for the true news cascade happens once at the beginning of the process in response to a detected fake news cascade, after which both cascades continue to propagate competitively without external moderation. Secondly, a user being ``activated" represents both exposure and re-sharing, since the model does not differentiate between them and assumes that an activated user is exposed and always attempts to activate its neighbors through forced re-sharing.

\subsubsection{Multi-stage intervention} \label{sec:hawkes}
Extension to a multi-stage intervention strategy was proposed by \citeauthor{pmlr-v70-farajtabar17a} \citeyear{pmlr-v70-farajtabar17a} to allow external interventions to adapt as necessary to the observed propagation dynamics of fake news. Specifically, \citeauthor{pmlr-v70-farajtabar17a} \citeyear{pmlr-v70-farajtabar17a} considered a different model of social influence based on multivariate point processes wherein past news sharing events trigger future news sharing events based on the intensities of influence and time delay between events. 

\paragraph{Approach} In multivariate point processes, each user $i$'s (news sharing) events are triggered by their base intensity $\mu_i$ and past (news sharing) events with some decay function $g$ that depends on the time gap between the current time and the past event, and the past events of other users $j$, whose influence is captured by $\alpha_{i,j}$. The intensity function of user $i$ is then $\lambda_i(t) = \mu_{i} + \sum_{t_j <t} \alpha_{i,j} g(t - t_j)$ and the probability of an event conditioned on past events is proportional to the intensity function. \cite{pmlr-v70-farajtabar17a} find that this model can naturally capture the diffusion of fake (true) news and allow one to separately capture being `exposed' to fake (true) news v/s being exposed and `sharing' fake (true) news, addressing the drawbacks of works mentioned in \ref{sec:compcas} which cannot make that distinction and assume that if a user is exposed to fake (true) news, the user also shares the fake (true) news with his/her social media connections. \citeauthor{pmlr-v70-farajtabar17a} \citeyear{pmlr-v70-farajtabar17a} model fake news with one point process $F(t)$ and its counteracting true (mitigation) news with another $M(t)$. $M(t)$ represents a vector with the number of times each user shares a mitigation event. $\mathcal{M}(t) = BM(t)$ where $B$ is the adjacency matrix represent event exposures, implying that a user is exposed whenever she or her neighbor shares the news. Similarly, fake news event shares and exposures counts are represented with $F(t)$ and $\mathcal{F}(t)$. The purpose of the intervention at any stage is to externally incentivize certain users (social network moderators can recommend, display or share the true event with them directly) in order to enable increased sharing of true news over the network that can counteract the fake news process. At each stage of the intervention certain budget and user activity constraints are imposed. The model allows us to solve for the optimal amount of external incentivization needed on every user at each stage under imposed constraints which will be enough to achieve a desired objective/reward from the intervention (such as minimizing the difference between fake and true news exposures).  \citeauthor{pmlr-v70-farajtabar17a} \citeyear{pmlr-v70-farajtabar17a} provide a reinforcement learning based policy iteration framework for optimization of the proposed multi-stage intervention objective to derive the optimal amount of external incentivization.

\paragraph{Limitation} The method assumes that one has already identified fake news and is tracking its propagation through the network, that is not trivial. Moreover, it could be arguable that once we have identified the fake news, we should simply use direct recommendations of true news to the users who we know are exposed to the fake news; and remove the fake news content from the platform to prevent future spread.

\subsection{Identification strategies}
As stated earlier in Section \ref{sec:intro}, studies on information cascades reveal that information is readily and rapidly transmitted over the network, even when it is of dubious
veracity \cite{friggeri2014rumor}. Given that even fake news can propagate exponentially quickly through reshares on social media, intervention efforts directed towards early identification and containment of fake news are of primary significance. In this section, we discuss works that suggest different mechanisms and intervention efforts to \emph{actively} detect and prevent the spread of fake news on social media. 

\subsubsection{Network monitoring} Intervention strategy based on network monitoring involves intercepting information from a list of suspected sources of fake news using computer-aided social media accounts or real paid user accounts, whose role is to filter the information they receive and block what they consider to be fake news. The network monitor placement can be determined by finding the cut or partition of the network which has the highest probability of transmission across the cut, and a maximum of $k$ users on the side of suspected sources \cite{amoruso2017contrasting}. Another possible network monitoring placement solution is based on a Stackelberg game between attacker and defender nodes \cite{yu2018adversarial}. The solution to the game allows the network administrator to select the set of nodes to monitor. Multiple monitoring sites with heterogeneous detection capabilities in both works are motivated by two ideas. One that having multiple check-points where fake news can be detected reduces the chance that the fake news would go undetected. Secondly, having multiple human or machine classifiers improves detection robustness, because something that is missed by one fact-checker might be captured by another.

\paragraph{Limitation} In changing network topologies, network monitoring solutions would require to update their strategy based on the change. Also, network monitoring on large-scale networks could be expensive due to the size of the network.

\subsubsection{Crowd-sourcing}
Another set of intervention efforts are designed to leverage crowd-sourcing mechanisms introduced on social media platforms which allow users to report or flag fake news articles. Instead of explicit network monitors as used by the approaches discussed earlier, these works implicitly monitor the network using crowd-sourced user feedback. However, a major drawback of the reporting or flagging mechanism is its possible misuse by adversaries. In addition, even signals from trustworthy users can be noisy since users are not all equally good at identifying misinformation. A data-driven assessment by \citeauthor{freeman2017can} \citeyear{freeman2017can} on data from \emph{LinkedIn}, suggests that only 1.3 \% of the users show measurable and repeatable skills at reporting actions performed by fake accounts. 

\paragraph{Approach} One way to leverage crowd-sourced signals to prioritize fact-checking of news articles was proposed in \citeauthor{kim2017leveraging} \citeyear{kim2017leveraging} by capturing the trade-off between the collection of evidence (flags) v/s the harm caused from more users being exposed to fake news (exposures) to determine when the news needs to be verified. They modeled the fact-checking process and events using point process models and derived the optimal intensity of fact-checking which is proportional to the rate of exposures to misinformation and the collected evidence as flags. In order to more accurately leverage user flags, \citeauthor{tschiatschek2017detecting} \citeyear{tschiatschek2017detecting} designed an online learning algorithm to jointly infer the flagging accuracies of users whilst identifying fake news. The algorithm selects $k$ news to inspect at each stage by greedily picking the ones that maximize the total expected utility - The total expected utility is an aggregation of the expected utility (reward) over each stage $t$ of the intervention, starting from stage 1, up to the last stage $T$ (that the algorithm is allowed to fix), and is mathematically defined as follows.
\[ \textrm{Util}(T, \textrm{ALGO})= \sum_{t=1}^T \mathbb{E} \left[ \sum_{s \in S^t} 1_{\{y^\star(s)=f\}} \textrm{val}^t(s) \right], \] 
where $\textrm{val}^t(s)$ is the expected reward at stage $t$ from the inspection and debunked of fake news from the news selected for fact-checking at stage $t$. The reward at stage $t$ can be modeled as the number of users prevented from seeing the fake news because of the intervention and debunking at stage $t$. \citeauthor{tschiatschek2017detecting} \citeyear{tschiatschek2017detecting} note that this quantity can be estimated by modeling and simulating the future spread to get the estimated number of users prevented from being exposed to the fake news that was debunked. $y^\star$ is the label of the article, estimated using Bayesian inference from $(\theta_{u,f}, \theta_{u, \bar{f}})$, which are parameters used to model the flagging accuracy of a user, and are estimated based on the history $D$ of fact-checked news by sampling from the posterior distributions $P(\theta_{u,f}|D)$ and $P(\theta_{u,\bar{f}}|D)$.

\paragraph{Limitation} Crowd-sourcing methods that essentially utilize reporting mechanisms, trade-off the number of users affected (exposed to fake news) with the number of reports (flags) needed to make a reliable prediction.

\subsubsection{User behavior modeling}

Apart from solutions based on explicit or implicit network monitoring, another proposed solution is based on a model of user behavior and news sharing \cite{papanastasiou2017fake}. The proposed model is developed to determine how to prioritize fact-checking of news articles by determining the optimal time of inspection of an article based on the proposed model of news sharing. The sharing of news is modeled as a sequential process among rational agents, wherein each agent chooses whether to verify the article, and chooses whether to share it with the next agent or not. By rational agents it is assumed that agents intend to share only true news. Under this model, the intervention can be designed as an optimal stopping problem solved using dynamic programming to find the optimal time of inspection. The model characterizes agent judgments in verifying fake news as well as the sequential interaction between agents, which are useful in determining the optimal time and priority of inspection. For instance, if the article is likely to be true or if is believed to be fake but more likely to be verified by the agents themselves before sharing, the need for inspection is lowered. 

\paragraph{Limitation} A drawback of the proposed method is that it is restricted to ``tree'' structured networks and cannot generalize to complex network topologies.
\\\\
\textbf{Limitations of existing intervention based methods.} Intervention based methods are more difficult to evaluate and test out, especially in complex environments like this, with lots of interdependent interactions. Also, they might make restrictive assumptions in certain cases which could limit their applicability.

\section{Existing Datasets} \label{sec:datasets}

The development of novel solutions to fake news detection has often been limited by data quality. In order to facilitate future research on this topic, we have compiled a comprehensive list of datasets related to the fake news detection task. Due to different research purposes, the data collections could vary significantly. For example, some datasets focus solely on political statements while others consist of open-domain news/articles. In addition, datasets could vary depending on what labels are provided, how are the labels collected, what types of text contents are included (e.g., source claims, user responses, etc.), whether temporal and propagation information is recorded, and the size of the collection. Therefore, we summarize these properties of the existing datasets. 






\subsection{Datasets enumeration}
The following is a comprehensive list of existing datasets. For named datasets, we use existing names as is; for unnamed ones, we assign suitable names for referencing.
\subsubsection{LIAR}
``Liar, Liar Pants on Fire": A New Benchmark Dataset for Fake News Detection. \cite{wang2017liar} ~\url{https://www.cs.ucsb.edu/̃william/data/liar_dataset.zip}

\subsubsection{Twitter}
Detecting Rumors from Microblogs with Recurrent Neural Networks. \cite{ma2016detecting} ~\url{http://alt.qcri.org/~wgao/data/rumdect.zip}

\subsubsection{Weibo}
Detecting Rumors from Microblogs with Recurrent Neural Networks. \cite{ma2016detecting} ~\url{http://alt.qcri.org/~wgao/data/rumdect.zip}

\subsubsection{FacebookHoax}
Some Like it Hoax: Automated Fake News Detection in Social Networks. \cite{tacchini2017some} ~\url{https://github.com/gabll/some-like-it-hoax/tree/master/dataset}

\subsubsection{PHEME-R}
Analyzing How People Orient to and Spread Rumours in Social Media by Looking at Conversational Threads. \cite{zubiaga_analysing_2016}
~\url{https://figshare.com/articles/PHEME_rumour_scheme_dataset_journalism_use_case/2068650}

\subsubsection{PHEME}
All-in-one: Multi-task Learning for Rumour Verification. \cite{colingPheme}
~\url{https://figshare.com/articles/PHEME_dataset_for_Rumour_Detection_and_Veracity_Classification/6392078}

\subsubsection{Cred-1}
Where the Truth Lies: Explaining the Credibility of Emerging Claims on the Web and Social Media. \cite{popat_where_2017}
~\url{http://www.mpi-inf.mpg.de/departments/databases-and-information\\-systems/research/impact/web-credibility-analysis/}

\subsubsection{Cred-2}
Where the Truth Lies: Explaining the Credibility of Emerging Claims on the Web and Social Media. \cite{popat_where_2017}
~\url{http://www.mpi-inf.mpg.de/departments/databases-and-information\\-systems/research/impact/web-credibility-analysis/}

\subsubsection{FakevsSatire}
Fake News vs Satire: A Dataset and Analysis \cite{golbeck2018fake}
~\url{https://github.com/jgolbeck/fakenews}

\subsubsection{BuzzfeedNews}
Hyperpartisan Facebook Pages Are Publishing False And Misleading Information At An Alarming Rate. \cite{silverman2016hyperpartisan} ~\url{https://github.com/BuzzFeedNews/2016-10-fac\\ebook-fact-check }

\subsubsection{KaggleFN}
Kaggle fake news dataset.~\url{https://www.kaggle.com/mrisdal/fake-news} 

\subsubsection{NewsFN}
GeorgeMcIntire/fake\_real\_news\_dataset~\url{https://github.com/GeorgeMcIntire/fake_real_news_dataset}

\subsubsection{BuzzfeedPolitical}
This Just In: Fake News Packs a Lot in Title, Uses Simpler, Repetitive Content in Text Body, More Similar to Satire than Real News. \cite{horne2017just}
~\url{https://github.com/BenjaminDHorne/fakenewsdata1}

\subsubsection{Political-1}
This Just In: Fake News Packs a Lot in Title, Uses Simpler, Repetitive Content in Text Body, More Similar to Satire than Real News. \cite{horne2017just}
~\url{https://github.com/BenjaminDHorne/fakenewsdata1}

\subsubsection{KaggleEmergent}
Kaggle rumors dataset.~\url{https://www.kaggle.com/arminehn/rumor-citation}

\subsubsection{NewsFN-2014}
Fact Checking: Task definition and dataset construction. \cite{vlachos2014fact}
~\url{https://sites.google.com/site/andreasvlachos/resources/FactChecking_LTCSS2014_release.tsv?attredirects=0 }

\subsubsection{NewsTrustData}
Leveraging Joint Interactions for Credibility Analysis in News Communities. \cite{mukherjee2015leveraging} ~\url{https://www.mpi-inf.mpg.de/departments/databases-and-\\information-systems/research/impact/credibilityanalysis/}

\subsubsection{FakeNewsNet-1}
FakeNewsNet: A Data Repository with News Content, Social Context and Dynamic Information for Studying Fake News on Social Media \cite{shu2018fakenewsnet}
~\url{https://github.com/KaiDMML/FakeNewsNet}

\subsubsection{FakeNewsNet-2}
FakeNewsNet: A Data Repository with News Content, Social Context and Dynamic Information for Studying Fake News on Social Media. \cite{shu2018fakenewsnet}
~\url{https://github.com/KaiDMML/FakeNewsNet}

\subsubsection{Twitter15}
Detect Rumors in Microblog Posts Using Propagation Structure via Kernel Learning. \cite{ma2017detect}
~\url{https://www.dropbox.com/s/7ewzdrbelpmrnxu/rumdetect2017.zip?dl=0}

\subsubsection{Twitter16}
Detect Rumors in Microblog Posts Using Propagation Structure via Kernel Learning. \cite{ma2017detect}
~\url{https://www.dropbox.com/s/7ewzdrbelpmrnxu/rumdetect2017.zip?dl=0}

\subsubsection{FEVER}
FEVER: A Large-scale Dataset for Fact Extraction and VERification. \cite{thorne_fever:_2018}
~\url{http://fever.ai/data.html}

\subsubsection{FNC-1}
Fake News Challenge: Stance Detection Dataset.~\url{https://github.com/FakeNews\\Challenge/fnc-1}

\subsection{Dataset Characteristics}
We provide a summarization of the characteristic features of listed datasets in Tables  \ref{tab:datasets_a},  \ref{tab:datasets_b}, and  \ref{tab:datasets_c} based on the task and application context, information content features and collected user responses features.

\begin{table*}
\centering
\caption{Datasets Characteristics. Task/Application and Annotation.} 
\label{tab:datasets_a}
\resizebox{0.9\textwidth}{!}{
\begin{tabular}{|l|p{3cm}|p{3cm}|p{2cm}|p{3cm}|}
\hline
\textbf{Dataset} & \textbf{Task/Application} & \textbf{Task labels} & \textbf{Content type} & \textbf{Annotator} \\ \hline
LIAR & fake news detection & pants-fire, false, barely-true, half-true, mostly-true, and true \footnotemark[1] & political statements & PolitiFact \\ \hline
FakevsSatire & fake news detection & fake, satire \footnotemark[1] & political news & researchers \\ \hline
NewsFN & fake news detection & fake, real & news articles & unspecified \\ \hline
BuzzfeedPolitical & fake news detection & fake, real & political news & verified list \\ \hline
Political-1 & fake news detection & fake, real, satire & political news & Zimdars websites list \\ \hline
NewsFN-2014  & fake news detection & true, mostly true, half true, mostly false, false \footnotemark[1] & fact-checked claims & Channel4 and PolitiFact \\ \hline
NewsTrustData & credibility assessment & qualitative scores \footnotemark[11] & news articles & NewsTrust member \\ \hline
Twitter & rumor classification & rumor, non-rumor & fact-checked claims & mapped from Snopes \footnotemark[5] \\ \hline
Weibo  & rumor classification & rumor, non-rumor & fact-checked claims & Sina community management center \\ \hline
Twitter15 & rumor classification &  rumor (false, true, remains unverified), non-rumor & fact-checked claims & mapped from Snopes \footnotemark[5] \\ \hline
Twitter16  & rumor classification & rumor (false, true, remains unverified), non-rumor & fact-checked claims & mapped from Snopes \footnotemark[5] \\ \hline
PolitiFact & fake news detection & fake, real & political statements & PolitiFact \\ \hline
GossipCop  & fake news detection & fake, real & entertainment news & GossipCop \\ \hline
FacebookHoax  & hoax detection & hoax, non-hoax & sicentific, conspiracy & non-hoax as scientific \\ \hline
PHEME-R \footnotemark[12] & rumor analysis  & rumor only (false, true, remains unverified) \footnotemark[2] & newsworthy stories & journalists\footnotemark[6], crowd-sourcing\footnotemark[7] \\ \hline
PHEME & rumor classification & rumor (false, true, remains unverified), non-rumor  & newsworthy stories & journalists\footnotemark[6], crowd-sourcing\footnotemark[7] \\ \hline
BuzzfeedNews & fake news detection & mostly true, mixture, mostly false, no factual content & political news & unspecfied \\ \hline
KaggleEmergent & rumor classification & rumor (false, true, unverified) & fact-checked claims & Emergent \\ \hline
KaggleFN & fake news detection & only fake & news articles & BS Detector \footnotemark[8] \\ \hline
Cred-1 & fact extraction & false, true & fact-checked claims & mapped from Snopes \footnotemark[9] \\ \hline
Cred-2 & fact extraction  & only false & wikipedia hoax & verified list \footnotemark[9] \\ \hline
FEVER & fact extraction & supported, refuted, not enough info \footnotemark[4] & constructed claims \footnotemark[3] & trained annotators \\ \hline
FNC-1 & stance detection & agrees, disagrees, discusses, unrelated & news articles & FakeNewsChallenge \footnotemark[10] \\ \hline
\end{tabular}
}
\\ \small\textsuperscript{1} with justifications \hspace*{0.2cm}
\small\textsuperscript{2} plus posts stance, certainty, evidentiality \hspace*{0.2cm}
\small\textsuperscript{3} using Wikipedia \hspace*{0.2cm}
\small\textsuperscript{4} with evidence \hspace*{0.2cm}
\small\textsuperscript{5} mapping policy not specified \hspace*{0.2cm}
\small\textsuperscript{6} for class label \hspace*{0.2cm}
\small\textsuperscript{7} other annotations \hspace*{0.2cm}
\small\textsuperscript{8} not fact-checked by humans \hspace*{0.2cm}
\small\textsuperscript{9} also provides Search Engine results for claims (but without annotating stance) \hspace*{0.2cm}
\small\textsuperscript{10} stance of the body towards the title \hspace*{0.2cm}
\small\textsuperscript{11} (objectivity, correctness, bias, credibility)  \hspace*{0.2cm}
\small\textsuperscript{12} used in RumorEval Task \hspace*{0.2cm}
\end{table*}

\begin{table*}
\caption{Dataset Characteristics. Information content to be classified (claims, articles, etc.).} \label{tab:datasets_b}
\resizebox{0.9\textwidth}{!}{
\begin{tabular}{|l| p{1cm} | p{2.6cm} | p{6.3cm}|}
\hline
\textbf{Dataset} & \textbf{Total claims}  & \textbf{Collection period} & \textbf{Number of claims (split by label)} \\ \hline
LIAR & 12.8K & 2007-2016 & roughly equal \\ \hline
FakevsSatire & 486 & 2016-2017 & 58\% fake \\ \hline
NewsFN & 6331 &   & roughly equal \\ \hline
BuzzfeedPolitical & 120 & Jan-Oct 2016 & roughly equal \\ \hline
Political-1 & 225 &  & roughly equal \\ \hline
NewsFN-2014 & 221 & 2013-2014 & 68\% half true, 50\% false, rest $\sim$35\% \\ \hline
NewsTrustData & 82K & per article & not applicable to qualitative scores \\ \hline
Twitter & 992 & Mar-Dec 2015 & roughly equal \\ \hline
Weibo & 4664 &  & roughly equal \\ \hline
Twitter15 & 1,490 & &  roughly equal \\ \hline
Twitter16 & 818 &  &  roughly equal \\ \hline
PolitiFact & 488 & dynamic & roughly equal \\ \hline
GossipCop & 3570 & dynamic & 19\% fake \\ \hline
FacebookHoax & 15.5K &  & 60\% hoax \\ \hline
PHEME-R & 330 &  & 50\% true, 20\% false, 30\% unverified \\ \hline
PHEME & 6425 & & 60\% non-rumor, rumor: 16\% true, 10\% false, 10\% unverified \\ \hline
BuzzfeedNews & 2282 &  19-27 Sep 2016 & 38\% mostly false / mixture \\ \hline
KaggleEmergent & 2145 &  Jan-Dec 2017 & 26\% false, 34\% true, 40\% unverified \\ \hline
KaggleFN & 13K & Oct-Nov 2016 & all fake \\ \hline
Cred-1 & 4856 &  & 74\% false \\ \hline
Cred-2 & 157 &  &  all false \\ \hline
FEVER & 185K &  & 55\% support, 20\% refute, 25\% other \\ \hline
FNC-1 & 50K & & 7\% agree, 2\% disagree, 18\% discuss \footnotemark[5] \\ \hline
\end{tabular}
}
\end{table*}



\begin{table*}
\centering{
\caption{Dataset Characteristics. Collected responses (user engagements) on social media. [y=yes, blank=no]. \label{tab:datasets_c}}
\resizebox{0.9\textwidth}{!}{
\begin{tabular}{| l | l | p{1cm} | p{1cm} | p{1cm} | p{0.8cm} | p{1.3cm} | l |}
\hline
\textbf{Dataset} & \textbf{Avg / claim} & \textbf{Time-stamp} & \textbf{Text} & \textbf{User info} & \textbf{Tree} & \textbf{Network} & \textbf{Platform} \\ \hline
Twitter & 1111 & y & y & y &  &  & Twitter \\ \hline
Weibo & 816 & y & y & y &  &  & Weibo \\ \hline
Twitter15 & 223 & y & y & y & y &  & Twitter \\ \hline
Twitter16 & 251 & y & y & y & y &  & Twitter \\ \hline
PolitiFact & 357 & y & y & y &  & y & Twitter \\ \hline
GossipCop & 10 & y & y & y &  & y & Twitter \\ \hline
FacebookHoax & 156 \footnotemark[1] & y & y & y &  &  & Facebook \\ \hline
PHEME-R & 15 & y & y & y & y & y & Twitter \\ \hline
PHEME & 16 & y & y & y & y &  & Twitter \\ \hline
BuzzfeedNews & 9792 \footnotemark[2] &  &  &  &  &  & Facebook \\ \hline
KaggleEmergent & 7187 \footnotemark[3] &  &  &  &  &  & not specified \\ \hline
KaggleFN & 27  \footnotemark[2] &  &  &  &  &  & Facebook \\ \hline
\end{tabular}
}
\\ \small\textsuperscript{1} only likes \hspace*{0.2cm}
\small\textsuperscript{2} only counts (shares, reactions, comments) \hspace*{0.2cm}
\small\textsuperscript{3}  only counts (shares) \hspace*{0.2cm}
}
\end{table*}


\subsubsection{Task and application context} In Table \ref{tab:datasets_a}, we mention the task/application context for each dataset along with the task labels provided in the dataset. We additionally mention the content types (domains) such as political news, entertainment news, Wikipedia statements, etc. that are encompassed by contents of each dataset. Lastly, we provide the annotation scheme used in the dataset which provides crucial insight into the data quality. For instance, KaggleFN dataset consists of news articles that are marked as fake by the \texttt{BS-Detector} application and not through journalistic or human verification, which means that the dataset labels could be noisy and moreover, training a model to learn those labels, is essentially training a model to essentially mimic \texttt{BS-Detector} \cite{shu2018fakenewsnet}.

\subsubsection{Information content} In Table \ref{tab:datasets_b} we provide additional details about the information content that are made available in each dataset. The information content (claim) to be verified can be an article, post, political statement, news stories, etc. as mentioned under Content Types in Table \ref{tab:datasets_a}. Here, we mention the total number of claims in each dataset. Total claims is the number of information contents that need to be verified. In addition we provide the division of claims by task labels such as the percentage of fake vs true articles in the dataset, which are useful for accessing applicability of machine learning models to a specific dataset. Lastly, we provide the collection period, that is the period in which the collected articles/claims were published or posted, which much like the domain (content types) is an important characteristic of the dataset. When the collection period is left blank, it means that it is unspecified in the dataset.

\subsubsection{User responses} Lastly, in Table \ref{tab:datasets_c} we provide features of user responses collected from user engagements with the information content, on social media. Note that this information is only provided in a subset of the datasets. We mention different features of the user responses in the Table such as average responses per claim, which is a popularly mentioned statistic in existing datasets. However, there can be high variation in the number of responses, based on virality of collected stories, and the platform used for collection such as Twitter, Weibo. The platform is also mentioned in the Table, for each dataset.\footnote{In accordance with the platform Twitter's information sharing policy, public release and distribution of tweet contents and metadata is restricted. Twitter based datasets generally release tweet ids and user ids from which complete information about contents and metadata is obtainable through Twitter Search API. For these datasets, for information that is obtainable from Twitter, we mark the information as available in our Tables regardless of whether it is released in the dataset directly or through the Twitter id information.} In addition, we note whether timestamp information, text information (in comments/replies), user information (user participating in the response), propagation tree information (who replies to whom) and social network information of users (follower-followee relationships) is provided in each dataset. Together, the three tables provide comprehensive insights into all existing datasets, facilitating a comparative analysis between them, in order to identify the potential shortcomings of existing datasets, as well as to identify the best choice of a dataset for a particular application/task, or to determine its suitability for training a specific detection model.

\section{Conclusion and Future Work}
The literature surveyed here has demonstrated significant advances in addressing the identification and mitigation of fake news. Nevertheless, there remain many challenges to overcome in practice.  We now outline three concrete directions for future research that can further advance the landscape of computational solutions.

\begin{enumerate}
\item
\textbf{Dynamic knowledge bases.} Although we can design many features which can potentially help us determine the truthfulness of a news article, the truthfulness is still ultimately defined by the statements it makes. The greatest challenge in developing automated fact-checking methods is the construction of dynamic knowledge bases, which can be regularly and automatically updated to reflect the changes occurring in a fast-paced world.\\

\item
\textbf{New intervention strategies.} For the design of useful intervention strategies for different environments, the most important question that needs to be answered is which environmental factors are most conducive to the spread of fake news. Studying and characterizing the relationship between user actions and utilities at the microscopic level of the individual, and the macroscopic impact in different networked environments, will be essential for explaining the spread of fake news and finding the best intervention strategies suited to that environment. Another interesting direction is towards \emph{educational} interventions. Recently, \citeauthor{gillani2018me} \citeyear{gillani2018me} studied the effect of educating people about the polarization of their social ego-networks using visualization tools that show users their social connections along with the inferred political orientations of other users in their network, to determine its impact on the user's connection diversity after the treatment. Another study proposed an \emph{active inoculation} strategy using a ``fake news game", where participants were asked to create content by using misleading tactics from the perspective of fake news creators \cite{roozenbeek2018fake}. The authors found preliminary evidence of its effectiveness in reducing perceived reliability and persuasiveness of fake news articles in a randomized field study of 95 high school students. \\

\item
\textbf{Datasets for intent detection.} Current datasets generally provide binary labels of information as fake or true. However, a more fine-grained classification of information by intent might be especially beneficial in identifying \emph{truly} fake news from closely related information such as satire and opinion news. In the list of fake websites maintained by \cite{zimdars2016false}, there is a \emph{type} label allowing up to 3 types for each website with tags such as political, satire, or biased. Some recent works have considered classification of fake vs satire news \cite{golbeck2018fake} and fake vs hyperpartisan news \cite{potthast2017stylometric} and we believe that this is an important direction for the future. 
\end{enumerate}


\begin{acks}
We thank the reviewers and moderators for their invaluable comments and inputs on earlier versions of this manuscript. This work is supported in part by the NSF Research Grant IIS-1619458 and IIS-1254206 as well as the National Natural Science Foundation of China NSFC Grant (NSFC Grant Nos.61772039, 91646202 and 61472006). The views and conclusions are those of the authors and should not be interpreted as representing the social policies of the funding agency, or the U.S. Government.
\end{acks}

\bibliographystyle{ACM-Reference-Format}
\bibliography{main}

\appendix
\section{Appendix}
We consolidate quantitative results on fake news (rumor) detection in terms of classification accuracy for several of the methods discussed earlier on a representative sample of the datasets namely Twitter collected by \cite{neuralurg2018}, Weibo and Twitter collected by \cite{ma2016detecting}, Twitter15 and Twitter16 collected by \cite{ma2017detect}; based on experimental evaluation performed in several works \cite{neuralurg2018,ma2018rumor,ma2016detecting,ruchansky2017csi}. While the list of methods and datasets is not exhaustive, there are other methods such as \citeauthor{ma2018detect} \citeyear{ma2018detect}, \citeauthor{kochkina18all} \citeyear{kochkina18all} and other datasets such as PHEME, PolitiFact, GossipCop \cite{kochkina18all,shu2018fakenewsnet} that could be evaluated, we regard a more extensive bench-marking based on specific application contexts with datasets of varying content types and collection periods as a direction for future work. Table \ref{tab:results_a}, \ref{tab:results_b} contain the consolidated quantitative results of classification accuracy for detection under content-based and feedback-based methods.

\begin{table*}
  \caption{Content-based (w/o user responses at test time) methods. Classification accuracy (\%). Results consolidated from \cite{neuralurg2018}.}
  \label{tab:results_a}
  \resizebox{0.75\textwidth}{!}{
  \begin{tabular} {lll}
    \toprule
   \textbf{Methods} & Weibo \cite{ma2016detecting}  & Twitter \cite{neuralurg2018} \\
    \midrule
    \citeauthor{ott2011finding} \citeyear{ott2011finding} - LIWC    & 66.06      & 62.13 \\ 
    \citeauthor{ott2011finding} \citeyear{ott2011finding} - POS    & 74.77      & 70.34 \\
    \citeauthor{ott2011finding} \citeyear{ott2011finding} - n-gram    & 84.76      & 80.69 \\
    \citeauthor{wang2017liar} \citeyear{wang2017liar}  & 86.23      & 83.24 \\
    \citeauthor{neuralurg2018} \citeyear{neuralurg2018}  & 89.84      & 88.83          \\
    \bottomrule
  \end{tabular}
  }
\end{table*}

\begin{table*}
\caption{{Feedback-based methods. Feature engineering, temporal, propagation. Classification accuracy (\%). Results consolidated from \cite{ruchansky2017csi,ma2018rumor}.} 
\label{tab:results_b}}
\resizebox{0.9\textwidth}{!}{
\begin{tabular}{llllll} 
\toprule
\textbf{Methods} & Weibo  & Twitter & Twitter15   & Twitter16\\
& \cite{ma2016detecting} & \cite{ma2016detecting} & \cite{ma2017detect} & \cite{ma2017detect} \\
\midrule
\citeauthor{zhao2015enquiring} \citeyear{zhao2015enquiring}                                                         & 73.2       & 64.4         & 40.9           & 41.4           \\ 
\citeauthor{castillo2011information} \citeyear{castillo2011information}                                                       & 83.1       & 73.1         & 45.4           & 46.5           \\ 
\citeauthor{kwon2017rumor} \citeyear{kwon2017rumor}                                                         & 84.9       & 77.2         & 56.5           & 58.5           \\ 
\citeauthor{ma2015detect} \citeyear{ma2015detect}                                                       & 85.7       & 80.8         & 54.4           & 57.4           \\ 
\citeauthor{ma2016detecting} \citeyear{ma2016detecting}                                                      & 91.1       & 88.1         & 64.1           & 63.3           \\ 
\citeauthor{ruchansky2017csi} \citeyear{ruchansky2017csi}                                                      & 95.3       & 89.2         & -              & -              \\ 
\citeauthor{wu2015false} \citeyear{wu2015false}                                                      & x          & x            & 49.3           & 51.1           \\ 
\citeauthor{ma2017detect} \citeyear{ma2017detect}                                                     & x          & x            & 66.7           & 66.2           \\ 
\citeauthor{ma2018rumor} \citeyear{ma2018rumor}                                                      & x          & x            & 72.3           & 73.7           \\ 
\bottomrule
\end{tabular}
}
\\ \small\textsuperscript{} x = Cannot be evaluated as per dataset features (information unavailable for method) \hspace*{0.2cm}
\end{table*}

\end{document}